\documentclass[lettersize,journal]{IEEEtran}
\usepackage{amsmath,amsfonts,amssymb}
\usepackage{algorithmic}
\usepackage{array}
\usepackage[caption=false,font=normalsize,labelfont=sf,textfont=sf]{subfig}
\usepackage{textcomp}
\usepackage{stfloats}
\usepackage{url}
\usepackage{verbatim}
\usepackage{graphicx}

\usepackage{times}
\usepackage{makecell}
\usepackage{soul}
\usepackage{colortbl}
\usepackage{multirow}
\usepackage{tablefootnote}
\usepackage{balance}
\usepackage{bm}
\usepackage[numbers]{natbib}
\usepackage{hyperref}
\usepackage{caption}
\newtheorem{definition}{Definition}

\newcommand{\rvc}[1]{{\color{black}#1}}

\newcommand{\gyx}[1]{{\color{black}#1}}
\newcommand{\rvg}[1]{{\color{black}#1}}
\newcommand{\cz}[1]{{\color{black}#1}}

\newcommand{\hy}[1]{{\color{black}#1}}

\def\BibTeX{{\rm B\kern-.05em{\sc i\kern-.025em b}\kern-.08em
    T\kern-.1667em\lower.7ex\hbox{E}\kern-.125emX}}
\usepackage{balance}
\begin{document}
\title{\rvc{Zero-shot and Few-shot} Learning with Knowledge Graphs: A Comprehensive Survey}
\author{Jiaoyan Chen, Yuxia Geng, Zhuo Chen, Jeff Z. Pan, Yuan He, Wen Zhang, Ian Horrocks and Huajun Chen
\thanks{Jiaoyan Chen (jiaoyan.chen@cs.ox.ac.uk), Yuan He and Ian Horrocks are from Department of Computer Science, University of Oxford, UK. Yuxia Geng, Zhuo Chen, Wen Zhang and Huajun Chen are from College of Computer Science and Technology, Zhejiang University, China. Jeff Z. Pan is from School of Informatics, The University of Edinburgh, UK.}
}

\markboth{Under revision}{}

\maketitle

\begin{abstract}
Machine learning especially deep neural networks have achieved great success but many of them often rely on a number of labeled samples for \rvc{supervision. As 
\rvg{sufficient labeled training data are not always ready}
due to e.g., \rvg{continuously} emerging prediction targets and costly sample annotation in real world applications, machine learning with sample shortage is now being widely investigated.
Among all these studies,
many prefer to utilize auxiliary information including those in the form of Knowledge Graph (KG)} 
to reduce the reliance on labeled samples.
In this survey, we \rvc{have} comprehensively reviewed over $\mathbf{90}$ papers about KG-aware research for two major \rvc{sample shortage} settings --- zero-shot learning (ZSL) \rvc{where some classes to be predicted have no labeled samples,}
and few-shot learning (FSL) where \rvc{some classes to be predicted} have only a small number of labeled samples that are available.
We first \rvc{introduce KGs} used in ZSL and FSL as well as their construction methods, and then systematically \rvc{categorize and summarize} KG-aware ZSL and FSL methods, dividing them into different paradigms such as the mapping-based, the data augmentation, the propagation-based and the optimization-based.
We next \rvc{present} different applications, including not only KG augmented prediction tasks \rvc{such as image classification, question answering, text classification and knowledge extraction, but also KG completion tasks, and some typical evaluation resources for each task.
We eventually discuss some challenges and open problems from different perspectives.}
\end{abstract}


\section{Introduction}

Machine learning (ML) especially deep learning is playing an increasingly important role in artificial intelligence (AI), and has achieved great success in many domains and applications in the past decades.
For example, Convolutional Neural Networks (CNNs) can often achieve even higher accuracy than human beings in image classification and visual object recognition, leading to the fast development of applications such as self-driving vehicles, face recognition, handwriting recognition, image retrieval and remote sensing image processing;
Recurrent Neural Networks (RNNs) and Transformer-based models are quite successful in sequence learning and natural language understanding, which boost applications such as machine translation, speech recognition and chatbots;
Graph Neural Networks (GNNs) have been widely applied to prediction tasks involving graph structured data in domains such as social networks, chemistry and biology.

However, the high performance of most ML models relies on a number of labeled samples for (semi-)supervised learning, while such labeled samples are often costly or not efficient enough to collect in real-word applications.
Even \rvc{when} labeled samples can be collected, re-training a complex model from scratch when new prediction targets (e.g., classification labels) emerge is unacceptable in many contexts where real-time is \rvg{required} or \rvc{enough computation \rvg{resource} is} inaccessible.
%
\rvc{All these situations will lead to \textit{sample shortage} in ML.
In the paper, we review two major sample shortage settings: \textit{zero-shot learning} (ZSL) and \textit{few-shot learning} (FSL).}
%
ZSL is formally defined as predicting new classes (labels) that have never appeared in training, where the new classes are  named as \textit{unseen classes} while the classes that have samples in training are named as \textit{seen classes} \cite{palatucci2009zero,lampert2009learning,farhadi2009describing}.
FSL is to predict new classes for which only a small number of labeled samples are given \cite{fink2005object}\cite{fei2006one}. 
For convenience, we also call such new classes with \rvg{insufficient} labeled samples as unseen classes, and the other classes that have a large number of samples used in training as seen classes.
Specially, when \rvc{the} unseen class has only one labeled sample, FSL becomes \textit{one-shot learning} \cite{fei2006one}.

ZSL has attracted wide attention in the past decade with quite a few solutions proposed \cite{fu2018recent,wang2019survey,chen2021knowledge}.
One common solution is transferring knowledge which could be samples, features (data representations) and model parameters from seen classes to unseen classes so as to \rvg{avoid only learning features from labeled samples} and training new models from the scratch \cite{pan2009survey}. 
For example, in zero-shot image classification, image features that have been already learned by CNNs such as ResNet 
from images of seen classes are often directly re-used to build classifiers for unseen classes.
The key challenge is selecting the right knowledge to transfer and adaptively combining these transferred knowledge for a new prediction task.
To this end, ZSL methods often utilize auxiliary information that \rvg{describes} inter-class relationships.
When ZSL was originally investigated for visual object recognition and image classification, the methods mainly use attributes that describe objects' visual characteristics (a.k.a. class attributes) \cite{lampert2009learning,farhadi2009describing}.
Next, class textual information such as class name and sentence description is widely studied due to its high accessibility \cite{frome2013devise,qiao2016less}.
In recent five years, \rvc{Knowledge Graph (KG), which often represents different kinds of knowledge such as relational facts, conceptualizations and meta information as RDF\footnote{Resource Description Framework, \url{https://www.w3.org/TR/rdf11-concepts/}} triples in form of $<$Subject, Predicate, Object$>$, has
attracted wide attention, and some KG-augmented ZSL methods have even achieved the state-of-the-art performance on many tasks \cite{wang2018zero,kampffmeyer2019rethinking,geng2021ontozsl}.
}

FSL, which started to attract wide attention around when one-shot learning was proposed \cite{fei2006one}, has a longer history and even more studies than ZSL \cite{wang2020generalizing}.
Since the unseen classes have some labeled samples although their sizes are quite small, techniques of \textit{meta learning} (a.k.a. \textit{learn to learn}) \cite{lemke2015metalearning} have been widely applied \cite{yin2020meta}.
Meta learning is usually applied by  either reducing the parameter searching space in training using meta parameters such as more optimized initial parameter settings, or transforming a classification problem to a metric learning problem where a testing sample is matched with the unseen classes based on their few-shot samples and meta learned mappings.
KGs have been utilized to optimise such meta learning-based methods; for example, Sui et al. \cite{sui2021knowledge} \rvc{retrieve} relevant knowledge from a KG named NELL \cite{mitchell2018never} to construct task-relevant relation networks as mapping functions for addressing few-shot text classification.
Meanwhile, the aforementioned idea of knowledge transfer can also be adopted for addressing FSL, where KG auxiliary information is becoming increasingly popular in recent years \cite{tsai2017improving,chen2018knowledge,peng2019few,zhang2020relation}.
For example, 
Chen et al. \cite{chen2018knowledge} \rvc{transfer} the feature learned by a CNN from flight delay forecasting tasks with a lot of historical records to a new forecasting task with \rvg{limited} historical records, by exploiting a KG with different kinds of flight related knowledge about e.g., airports and airlines;
Peng et al. \cite{peng2019few} \rvc{extract} a KG from WordNet for representing class hierarchies and then used this KG to augment knowledge transfer for few-shot image classification.

\vspace{0.1cm}
\noindent\textbf{Motivation and Contribution.} 
Since KG has become a very popular form for representing knowledge and graph structured data, acting as the foundation of many successful AI and information systems  \cite{hogan2020knowledge}, it is quite reasonable to use KGs to augment both ZSL and FSL as discussed above.
Quite a few papers have been published on KG-aware \rvc{zero-shot and few-shot} learning especially in recent five years, and this research topic is becoming more and more popular.
\rvc{It is worth mentioning that this topic
includes not only using KGs to augment ZSL and FSL but also addressing prediction tasks of the KG itself where ZSL and FSL methods are applied and extended for the KG context.}
%
By the middle of December in 2021, we have collected $50$ papers on KG-aware ZSL and $46$ papers on KG-aware FSL. 
To systematically categorize and compare all the proposed methods, and to present an overall picture of this promising field, a comprehensive survey is now in urgent need.
In this paper, we \textit{(i)} \rvc{introduce} KGs and their construction methods for \rvc{ZSL and FSL, \textit{(ii) categorize, analyze and compare} different kinds of KG-aware ZSL and FSL methods (see Figure \ref{fig:categories} for an overview of the paradigms and categories), \textit{(iii)} present ZSL and FSL} tasks as well as their evaluation resources \rvg{in various} domains including computer vision (CV), NLP and KG completion, and \textit{(iv)} \rvc{discuss} the existing challenges and potential future directions.
This survey is suitable for all AI researchers, especially those who are to enter the domain of \rvc{ML with sample shortage, those who have already been working on this topic but are interested in solutions utilizing} knowledge representation and reasoning,
and those who are working on KG and semantic techniques.

\rvc{
\begin{figure*}
\centering
\includegraphics[width=0.85\textwidth]{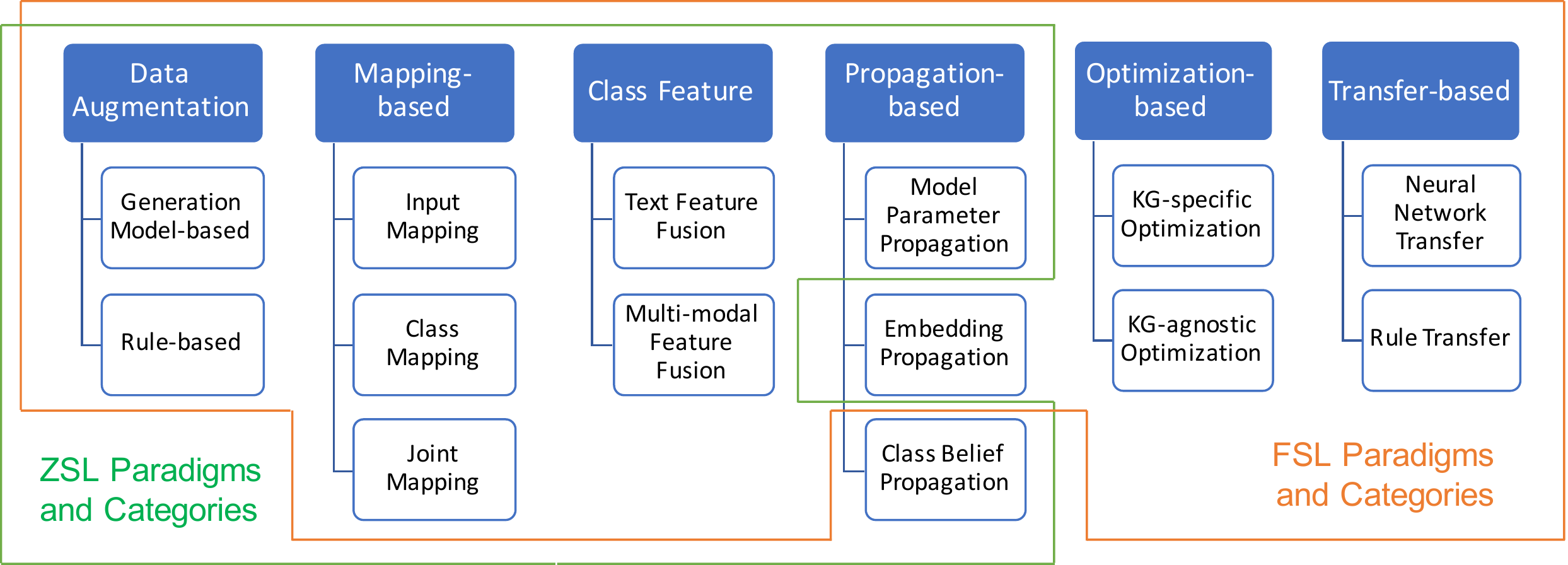}
\caption{Paradigms and categories of KG-aware zero-shot and few-shot learning methods \label{fig:categories}}
\end{figure*}
}

\vspace{0.1cm}
\noindent\textbf{Related Literature Reviews.} There have been several papers that have literature reviews relevant to \rvc{zero-shot and few-shot} learning, but they are all quite different from this survey.
\begin{itemize}
\item  The two survey papers \cite{wang2019survey} and \cite{wang2020generalizing} systematically review the ZSL methods by 2019 and the FSL methods by 2020, respectively, mainly from the perspective of problem setting (e.g., whether the unlabeled testing samples are used or not in training), ML theory (e.g., which prediction error to reduce), and methodology (e.g., data focused, model focused and learning algorithm focused). However, they do not consider the categorization and deep analysis from the perspective of auxiliary information, and failed to collect most KG-aware methods.

\item  The very recently released paper \cite{hu2021can} reviews both ZSL and FSL methods that use or aim at structured data. Structured data, however, is more general than KG with a much larger scope, and thus \cite{hu2021can} collects only a small part of the papers on KG-aware ZSL and FSL research. It includes $19$ papers about KG-aware ZSL and $21$ papers about KG-aware FSL, while this survey has $50$ papers and $46$ papers, accordingly.
This survey also has a more fine-grained method categorization, and additional technical analysis on KGs and their construction for \rvc{zero-shot and few-shot} learning. Meanwhile, \cite{hu2021can} focuses more on addressing problems in structured data by ZSL and FSL methods, but less on augmenting ZSL and FSL methods.

\item The paper \cite{chen2021knowledge} is our previous survey and perspective paper published in IJCAI 2021 Survey Track. It briefly categorizes different external knowledge used in ZSL with incomplete reviews on KG-aware ZSL papers, and it does not cover FSL.

\item The benchmarking paper \cite{xian2018zero} was published in 2018. It reviews around $10$ ZSL methods that mainly utilize class attribute and text information as the auxiliary information, focusing on their evaluation and result comparison on image classification task. This paper covers neither state-of-the-art ZSL methods proposed in recent $3$ years nor KG-aware ZSL methods. Similarly, the survey paper \cite{fu2018recent} reviews ZSL papers published before 2018, mainly focusing on ZSL studies on CV tasks.
\end{itemize}

\vspace{0.1cm}
\noindent\textbf{Paper Organization.}
The remainder of this survey is organized as follows. 
Section \ref{sec:preliminary} introduces the preliminary, including the \rvc{definitions and annotations of ZSL and FSL}, and an overall view of the auxiliary information.
Section \ref{sec:kg} introduces the definition and scope of KGs, as well as \rvc{the KG construction \rvg{for} ZSL and FSL}.
Section \ref{sec:zsl} reviews KG-aware ZSL methods which are categorized into four paradigms: mapping-based, data augmentation, knowledge propagation and feature fusions. For each paradigm, we further introduce different categories and their corresponding methods.
Section \ref{sec:fsl} is similar to Section \ref{sec:zsl} but reviews KG-ware FSL methods, \rvc{and compares KG-aware FSL and ZSL in the end.}
Section \ref{sec:app} introduces the development and resources of KG-aware ZSL and FSL in different tasks across CV, NLP and KG completion.
Section \ref{sec:challenge} discusses the existing challenges and the future directions.
Section \ref{sec:conclusion} concludes this paper.

\section{Preliminary on \rvc{Zero-shot and Few-shot} Learning}\label{sec:preliminary}

\rvc{
Both ZSL and FSL have been applied in many different tasks, varying from image classification and visual question answering to text classification, knowledge extraction and KG completion.
Although the exact ZSL and FSL problems may be different between papers, they can be expressed under one framework.
In the session, we aim to present this framework with formal problem definitions and annotations, and at the same time introduce some background knowledge that are needed for understanding KG-aware ZSL and FSL.
We start from ZSL, and then introduce FSL based on ZSL.
}

\begin{table}[t]
\footnotesize{
\centering
\renewcommand{\arraystretch}{1.3}
\begin{tabular}[t]{m{2cm}<{\centering}|m{5.8cm}<{\centering}}\hline
 \textbf{Annotation} &\textbf{Introduction}  \\ \hline
$\mathcal{D}_{tr}$, $\mathcal{X}_{s}$, $\mathcal{Y}_{s}$  & Training sample set, inputs of the training samples, seen classes, respectively \\ \hline
$\mathcal{D}_{te}$, $\mathcal{X}_{u}$, $\mathcal{Y}_{u}$  & Testing sample set, inputs of the testing samples, unseen classes, respectively \\ \hline
$\mathcal{D}_{few}$, $\mathcal{X}_{few}$  & Few-shot sample set, inputs of the few-shot samples, respectively \\ \hline
$f:x\rightarrow y$ & The target function mapping the input $x$ to the output class $y$ \\ \hline
$f':(x,y)\rightarrow s$ & The function that scores the matching degree between $x$ and $y$ \\ \hline
$g:x\rightarrow \bm{x}$ & The encoding function of the input $x$ \\ \hline
$h:y\rightarrow \bm{y}$ & The encoding function of the class $y$ \\ \hline

\end{tabular}
\caption{\rvc{A list of the ZSL and FSL annotations in the paper}.}\label{table:zsl}
}
\end{table}

\subsection{Zero-shot Learning}

\rvc{We first give a simple but generic definition towards ML classification, then formally define ZSL and introduce auxiliary information, and finally introduce the existing categorization of ZSL works. 
\begin{definition}[\rvg{Supervised} ML Classification]
\rvg{Given a set of labeled training samples $\mathcal{D}_{tr} = \{(x, y) | x \in \mathcal{X}, y \in \mathcal{Y}\}$, a classifier is trained to approximate a target function from the input $x$ to the output label $y$, denoted as $f:x\rightarrow y$, such that $f$ is able to correctly predict the labels of samples in a testing set $\mathcal{D}_{te} = \{(x, y) | x \in \mathcal{X}', y \in \mathcal{Y}\}$, where $\mathcal{X} \cap \mathcal{X}' = \emptyset$.
}
\end{definition}
}

In image or text classification, \rvc{$x$ is an image or text while $y$ is \rvg{an image category or a text category}.}
Sometimes, one input can be annotated by multiple labels, which is known as multi-label classification.
\rvc{In question answering}, we refer to giving an answer or multiple answers to a natural language question w.r.t. a given textual context, where \rvg{the label $y$ corresponds to the answer}.
Visual question answering is similar but the context is an image or a video.
\rvc{In knowledge extraction, the task is usually to extract entities, relations or events from natural language text. 
It also includes entity or relation linking, which matches an entity or relation mention in text with a pre-defined entity or relation, and entity typing, which assigns a pre-defined class or multiple predefined classes to an entity mention in the text.
Thus $x$ is often a sentence or a document with an entity or relation mention, while $y$ is a label corresponding to an entity, a relation, an event, or a class.
In KG completion, which is to predict a missing RDF triple, $x$ is often the two components of a triple while $y$ is a label indicating the third component.
Note in all these tasks, a candidate set is usually given for the output class $y$.}

\rvg{In supervised ML classification, given $\mathcal{D}_{tr}$, the trained classifier can only predict samples of classes that have appeared in the training stage (i.e., $\mathcal{Y}$), while ZSL aims to predict samples beyond $\mathcal{Y}$.}
\rvc{Here is the formal definition.
\begin{definition}[Zero-shot Learning]\label{def:zsl}
\rvg{
Given a training sample set $\mathcal{D}_{tr} = \{(x, y) | x \in \mathcal{X}_{s}, y \in \mathcal{Y}_s\}$ where $\mathcal{X}_{s}$ and $\mathcal{Y}_s$ are the training sample inputs and their classes, respectively,
\textit{Standard ZSL} aims to train $f$ with $\mathcal{D}_{tr}$ for predicting on a testing set $\mathcal{D}_{te} = \{(x, y) | x \in \mathcal{X}_u, y \in \mathcal{Y}_u\}$ where $\mathcal{X}_{u}$ and $\mathcal{Y}_u$ are the testing sample inputs and their classes, respectively, with $\mathcal{Y}_u \cap \mathcal{Y}_s = \emptyset$.
$\mathcal{Y}_s$ is called \textit{seen classes} while $\mathcal{Y}_u$ is called \textit{unseen classes}.}
When \rvg{it is required to predict testing samples of both seen and unseen classes}, i.e., $\mathcal{D}_{te} = \{(x, y) | x \in \mathcal{X}_u \rvg{\cup \mathcal{X}'_s}, y \in \mathcal{Y}_u \cup \mathcal{Y}_s\}$ with $\mathcal{X}_s \cap \mathcal{X}'_s = \emptyset$, the problem becomes \textit{Generalized ZSL}.
\end{definition}
}

\rvc{It is worth mentioning that in addressing some ZSL tasks such as text classification and question answering, the original function $f$ is sometimes transformed into a new scoring function by moving $y$ to the input, denoted as $f':(x,y)\rightarrow s$, where the output $s$ is a score indicating whether $y$ \rvg{is the label of} $x$ or not.
The label of a testing sample $x$ in $\mathcal{X}_u$ \rvg{(resp. $\mathcal{X}_u \cup \mathcal{X}'_s$)} is predicted by finding out the class in $\mathcal{Y}_u$ (resp. $\mathcal{Y}_u \cup \mathcal{Y}_s$) that maximizes the score $s$. }

\rvc{
\begin{definition}[Auxiliary Information]
\textit{Auxiliary information} is some kind of symbolic data that describe or indicate the relationship between seen and unseen classes, such as class attribute, class text description and class hierarchy.
With the auxiliary information, classes are usually encoded into sub-symbolic representations (i.e., vectors) with the relationship between classes concerned in the vector space.
We denote the class encoding as the function $h: y \rightarrow \bm{y}$ where the bolded $\bm{y}$ represents the vector of the class $y$, $y \in \mathcal{Y}_u \cup \mathcal{Y}_s$.\footnote{
The raw input $x$ could also be encoded by e.g., some pre-trained models or hand-craft rules before they are fed to $f$ (or $f'$). This step is optional but is often adopted.
We denote this initial encoding function as $g: x \rightarrow \bm{x}$ where the bolded $\bm{x}$ represents the encoding vector of $x$, $x\in\mathcal{X}_s\cup\mathcal{X}_u$.}
\end{definition}
}

\rvc{
Since unseen classes have no labeled samples, ZSL methods rely on \textit{auxiliary information}. 
In early years when ZSL was proposed in around 2009 for image classification, the majority of the solutions utilize class attributes which are often a set of key-value pairs for describing object visual characteristics \cite{lampert2009learning,lampert2013attribute}.
There are also relative attributes which enable comparing the degree of each attribute across classes (e.g., ``bears are furrier than giraffes'') \cite{parikh2011relative}, and real valued attributes for quantifying the degrees \cite{lampert2013attribute,xian2018zero}. 
The advantages and disadvantages of the attribute auxiliary information are quite obvious: it is easy to use and quite accurate with little noise, but it cannot express complex semantics for some tasks and is not easily accessible, usually requiring annotation by human beings or even domain experts.
From around 2013, class textual information, varying from words and phrases such as class names to long text such as sentences and documents for describing classes, started to attract wide attention in ZSL.
Typical works lie in not only image classification  \cite{frome2013devise,elhoseiny2013write} but also other tasks such as KG completion \cite{qin2020generative}.
Text information is easy to access for common ZSL tasks. It can be extracted from not only the data of the ZSL tasks themselves but also encyclopedias, Web pages and other online resources. 
However, it is often noisy with irrelevant words and the words are \rvg{often} ambiguous, failing to accurately express fine-grained, quantified or more complex inter-class relationships.}

In recent years, 
structured knowledge 
in the scope of KG, such as \rvc{class hierarchies and commonsense knowledge}, have become \rvc{increasingly} popular in ZSL research with very promising performance achieved.
Such knowledge can often express richer semantics than attributes and text, even including logical relationships, and at the same time, they become more available with the development of KG construction techniques and the availability of many public KGs such as WordNet \cite{miller1995wordnet}, ConceptNet \cite{speer2017conceptnet} and Wikidata \cite{vrandevcic2014wikidata}.
In this survey, we mainly review KG-aware ZSL studies, \rvc{using Section \ref{sec:kg} to independently introduce the involved KGs, and Section \ref{sec:zsl} to review the involved} methods.

\rvc{The survey paper \cite{wang2019survey} has categorized ZSL methods into} the following two categories:
\begin{itemize}
\item \textit{Classifier-based}. The classifier-based methods are to directly learn a classifier for each unseen class. 
They could be further divided into \textit{(i)} \textit{Corresponding Methods} which exploit the correspondence between the binary one-vs-rest classifier for each class and its corresponding encoding of the auxiliary information, \textit{(ii)} \textit{Relationship Methods} which calculate and utilize the relationships among classes, and \textit{(iii)} \textit{Combination Methods} which combine classifiers for basic elements that are used to constitute the classes.
\item \textit{Instance-based}. The instance-based methods are to obtain labeled samples belonging to the unseen classes and use them for learning and prediction.
They are further divided into \textit{(i)} \textit{Projection Methods} which learns a function to project the input and the class encoding into the same space (i.e., the class encodings after projection are regarded as labeled samples), \textit{(ii)} \textit{Instance-borrowing Methods} which transfer samples from seen classes to unseen classes, and \textit{(iii)} \textit{Synthesizing Methods} which obtain labeled samples for the unseen classes by synthesizing some pseudo samples.
\end{itemize}

This categorization is mainly from the perspective of ML theory and method.
It aims at general ZSL methods, no matter what kind of auxiliary information is utilized.
In contrast, our categorization which is to be introduced in Section \ref{sec:zsl} is from the perspective of auxiliary information, and focuses on more fine-grained comparison and analysis towards those KG-aware ZSL methods.
Meanwhile, since the survey \cite{wang2019survey} was published in 2019 while many KG-aware ZSL methods were proposed in recent two years, the KG-aware methods \rvc{covered are not complete}.

\subsection{Few-shot Learning}

\rvc{We first formally define FSL, following the annotations in defining ZSL, then introduce the auxiliary information and finally present the current method categorization.}

\rvc{
\begin{definition}[Few-shot Learning]
Given 
a set of training samples $\mathcal{D}_{tr} = \left\{ (x,y) | x \in \mathcal{X}_{s}, y \in \mathcal{Y}_s \right\}$ and a set of few-shot samples $\mathcal{D}_{few} = \left\{ (x,y) | x \in \mathcal{X}_{few}, y \in \mathcal{Y}_u \right\}$, 
where $\mathcal{Y}_u \cap \mathcal{Y}_s = \emptyset$, \rvg{each class in $\mathcal{Y}_{s}$ has a large number of samples in $\mathcal{D}_{tr}$ and} each class in $\mathcal{Y}_{u}$ has only a small number of samples in $\mathcal{D}_{few}$,
FSL is to train a classifier $f$ with $\mathcal{D}_{tr}$ and $\mathcal{D}_{few}$ for predicting samples in a testing set $\mathcal{D}_{te} = \left\{ (x,y) | x \in \mathcal{X}_{u}, y \in \mathcal{Y}_u \right\}$ with $\mathcal{X}_u \cap \mathcal{X}_{few} = \emptyset$, \rvg{or $\mathcal{D}_{te} = \left\{ (x,y) | x \in \mathcal{X}_{u} \cup \mathcal{X}'_{s}, y \in \mathcal{Y}_u \cup \mathcal{Y}_s \right\}$} with $\mathcal{X}_s \cap \mathcal{X}'_s = \emptyset$.
\end{definition}
}

\rvc{To be consistent with ZSL, we keep calling the classes with a large number of training samples, i.e., $\mathcal{Y}_s$, as \textit{seen classes}, those classes with few-shot labeled samples in $\mathcal{D}_{few}$, i.e., $\mathcal{Y}_u$, as \textit{unseen classes}.
As in ZSL, the original target of learning $f$ can also be transformed into learning a scoring function for ranking the candidate classes, i.e., $f': (x,y) \rightarrow s $.
}

\begin{figure}
\centering
\includegraphics[width=0.45\textwidth]{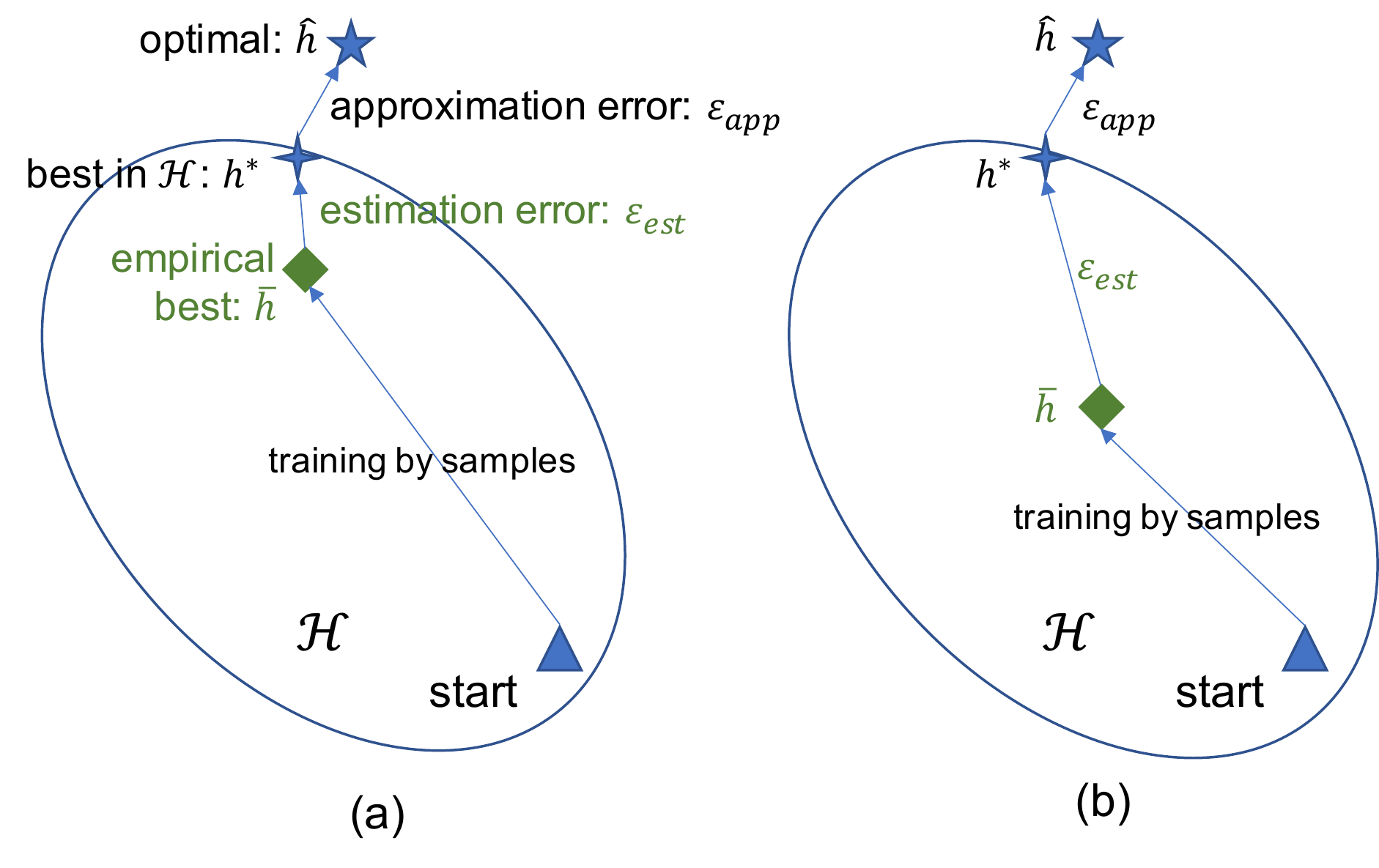}
\caption{Expected risk with (a) sufficient samples and (b) limited samples for training \cite{wang2020generalizing}. \label{fig:risk}}
\end{figure}

\rvc{The few-shot samples samples in $\mathcal{D}_{few}$ can be just one labeled sample per class, which is known as \textit{one-shot learning}. 
It can also have multiple labeled samples, but the size is relative small and they alone are far from enough to train a robust model for an unseen class.}
To be more specific, we introduce the concept of \textit{expected risk} as in \cite{wang2020generalizing}.
For an optimal hypothesis $\hat{h}$ (i.e., the target function $f$), its expected risk is composed of two parts: \textit{(i)} approximation error $\mathcal{E}_{app}$ which measures how close the best hypothesis $h^*$ in a given hypothesis set $\mathcal{H}$ can approximate $\hat{h}$, 
and \textit{(ii)} estimation error $\mathcal{E}_{est}$ which measures the effect of minimizing the empirical risk of the learned hypothesis $\bar{h}$ w.r.t. the best hypothesis $h^*$ \cite{bottou2018optimization}.
As shown in Figure \ref{fig:risk}, \rvc{model training for unseen classes with $\mathcal{D}_{few}$ has a much higher estimation error than model training for seen classes with $\mathcal{D}_{tr}$.}

\rvc{
The key difference of FSL in comparison with ZSL lies in the few-shot samples $\mathcal{D}_{few}$.
Most FSL methods now focus on fully utilizing $\mathcal{D}_{few}$.
They prefer some ML algorithms such as multi-task learning which allows parameter sharing between tasks, meta learning which directly predicts some parameters and hyper-parameters that are to learn or to adjust, and metric learning which compares a testing sample with the few-shot samples of each unseen class in some space after projection.

The common aspect of FSL and ZSL lie in the utilization of the auxiliary information.
All the auxiliary information used in ZSL can also be used in FSL, and the utilization method can be transferred to FSL easily. 
Even the few-shot samples can be regarded as an additional kind of auxiliary information.
KG has also been investigated in FSL as effective auxiliary information.
In FSL, we also denote the encoding of the class with the auxiliary information as $h: y \rightarrow \bm{y}$.
}
%
%

\rvc{The survey \cite{wang2020generalizing} divides FSL methods into the three general categories according to the aspects that are augmented:}
\begin{itemize}
\item \rvc{Data augmentation methods}. They increase the size of the few-shot samples ($\mathcal{D}_{few}$) via data augmentation by e.g., transforming samples from the training set $\mathcal{D}_{tr}$ \rvc{or other similar labeled data}, and generating samples from weekly labeled or unlabeled data.
\item \rvc{Model augmentation methods}. They reduce the original hypothesis set $\mathcal{H}$ to a small one for reducing the searching space. They \rvc{are} further divided into \textit{(i)} \rvc{multi-task learning methods which share parameters between tasks or to regularize} the parameters of the target task, \textit{(ii)} \rvc{embedding methods which} project samples to an embedding space where similar and dissimilar samples can be easily discriminated, \textit{(iii)} \rvc{generative modeling methods which restrict} the model distribution, and so on.
\item \rvc{Algorithm augmentation methods}. They guide and accelerate the searching of the parameters of the best hypothesis $h^*$ by e.g., learning the optimizer and aggregating existing parameters.
\end{itemize} 
\rvc{Although this is a systematic categorization, it has a limited} coverage on KG-aware FSL methods, and ignores the role of the auxiliary information especially KGs.
In this survey, we categorize and compare KG-aware FSL methods from the perspective of how KG is exploited. 

\section{Knowledge Graph} \label{sec:kg}

\rvc{In this section we will first introduce the definition and different forms of KG, and then present the existing KGs that have been adopted in ZSL and FSL studies, as well as KG construction methods for specific ZSL or FSL tasks. See Figure \ref{fig:kg} for an overview.}

\rvc{
\begin{figure*}
\centering
\includegraphics[width=0.92\textwidth]{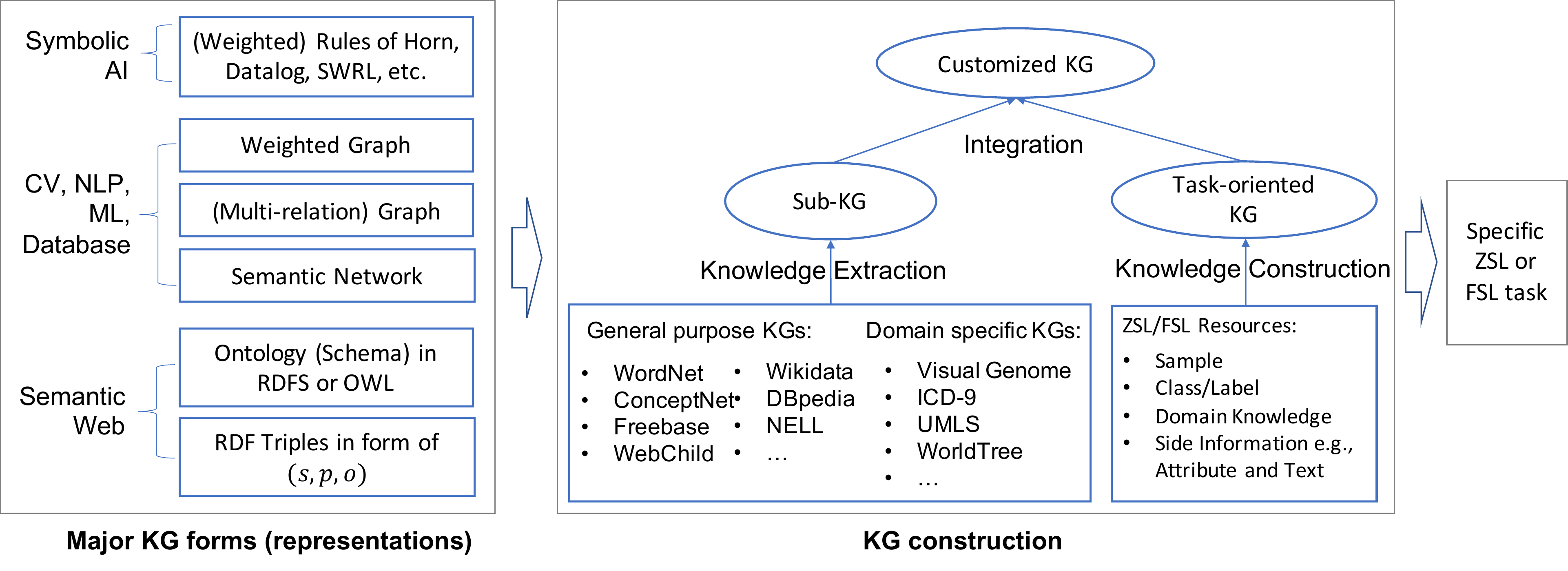}
\caption{\rvc{An overall picture of the KGs in ZSL and FSL} \label{fig:kg}}
\end{figure*}
}

\subsection{Definition and Scope}

Knowledge Graph (KG) \rvc{has been} widely used for representing graph structured knowledge, and has achieved great success in many \rvc{applications such as search engine, recommendation system, clinic AI and personal assistant} \cite{Pan2016,hogan2020knowledge}. 
In this \rvc{session we first introduce a basic but widely recognized KG definition and some basic KG access operations, then introduce the ontology-equipped KG from the semantic Web perspective, and finally introduce the scopes of KG in domains beyond the semantic Web.}

\rvc{
\begin{definition}[Knowledge Graph]
A KG, denoted as $\mathcal{G} = \left\{E, P, L, T\right\}$, is composed of an entity set $E$, a property set $P$, a literal set $L$, and a statement set $T$ in the form of RDF triple.
Each RDF triple in $T$ is denoted as ($s, p, o$), where $s$ represents the subject which is an entity in $E$, $p$ represents a predicate which is a property in $P$, and $o$ represents the object which can be either an entity in $E$ or a literal in $L$.
\end{definition}
}
\rvc{Some statements represent} relational facts. In this case, $o$ is also an entity, and $p$ is a relation between two entities (a.k.a., object property).
$s$ and $o$ are also known as the head entity and tail entity, respectively.
A set of relational facts composes a multi-relational graph whose nodes correspond to entities and edges are labeled by relations.
\rvc{Some other statements} represent literals as e.g., entity attributes. In this case, the predicate $p$ uses a data property and $o$ is a literal with some data type such as string, date, integer and decimal.
The literals also include KG meta information such as entity's label, textual definition and comment, which are also represented via built-in or bespoke annotation properties. 

\rvc{
The content of a KG can usually be efficiently accessed via two operations: \textit{lookup} and \textit{query}. 
KG lookup (a.k.a. KG retrieval) is a service that returns the most relevant entities  and/or properties in a KG that match the meaning of an input string (usually a phrase). For fast retrieval, some lexical index is usually built based on the labels and other name information of the entities and relations. 
%
KG query is a service that returns the answers of an input query of the RDF query language SPARQL\footnote{\url{https://www.w3.org/TR/rdf-sparql-query/}}. 
The input of such a query is actually a sub-graph pattern with variables, while the output could be not only the matched entities, properties and/or literals, but also the whole sub-graphs (i.e., triples).
Some modern graph databases such as RDFox \cite{nenov2015rdfox} can already support efficient SPARQL query.
}

\rvc{In the semantic Web, a KG} is often accompanied by an ontology as the schema, using languages from the semantic Web community such as RDFS\footnote{RDF Schema, \url{https://www.w3.org/TR/rdf-schema/}} and OWL\footnote{Web Ontology Language, \url{https://www.w3.org/TR/owl2-overview/}} for richer semantics and higher quality \cite{horrocks2008ontologies}. 
They often define hierarchical classes (a.k.a. concepts)\footnote{\rvc{To distinguish \textit{class} in ML and \textit{class} in KG, we use \textit{concept} for the latter.}}, properties (i.e., stating the terms used as relations), concept and relation hierarchies, constraints (e.g., relation domain and range, and class disjointness), and logical expressions such as relation composition.
The languages such as RDF, RDFS and OWL have defined a number of built-in vocabularies for representing these knowledge, such as \textit{rdf:type}, \textit{rdfs:subClassOf}, \textit{owl:disjointWith} and \textit{owl:someValuesFrom}. 
Note RDFS also includes some built-in annotation properties such as \textit{rdfs:label} and \textit{rdfs:comment} for defining the above mentioned meta information. 
With these vocabularies, an ontology can be represented as RDF triples; for example, the subsumption between two concepts can be represented by the predicate \textit{rdfs:subClassOf}, while the membership between an instance and a concept can be represented by the predicate \textit{rdf:type}.
The ontology alone, which is widely used to define domain knowledge, conceptualization and vocabularies such as terms and taxonomies, is also widely recognized as a KG.
One typical example is SNOMED CT which systematically organizes medical terms as entities with names, definitions, existential restrictions, tree-like categorizations and so on \cite{schulz2009snomed}. 
It is worth mentioning that KGs, especially those OWL ontologies and those equipped with ontologies, can support symbolic reasoning, such as consistency checking which can find logical violations, and entailment reasoning which infers hidden knowledge according to Description Logics. 

Besides the relational facts, literals and ontologies defined following \rvc{the semantic Web standards}, we also regard graph structured knowledge in some other forms as KGs, according to the terminologies and definitions used in other domains including ML, \rvc{Database}, CV and NLP.
One popular KG form is Semantic Network which can be understood as a graph that connects different concepts (entities) often with labeled edges for representing different relationships. 
Two such KGs that are widely used in many domains are WordNet which is a lexical database with different relationships between words \cite{miller1995wordnet} and ConceptNet which stores commonsense knowledge and relationships between different terms \cite{speer2017conceptnet}. 
We further relax the scope of KGs to single relation graphs such as simple taxonomy (i.e., a set of hierarchical classes) and graphs with weighted edges which may represent some quantitative relationships such as similarity and distance between entities.

\rvc{We} also regard logical rules of different forms, such as Horn clause, Datalog rules and SWRL\footnote{Semantic Web Rule Language, \url{https://www.w3.org/Submission/SWRL/}} rules, as well as their soft or fuzzy extensions (i.e., weighted rules) \cite{PSTH05}, within the scope of KG.
This is because many of these rules can be transformed into equivalent relational facts and ontological knowledge, and vice versa \cite{horrocks2004proposal}.
They can often be understood as logic models over KGs, through which hidden knowledge can be inferred.

\rvc{
\subsection{General Purpose KGs}
There have been several general purpose large-scale KGs that are open and can be directly utilized for different kinds of tasks. In this part we introduce these KGs and briefly present how their knowledge are extracted for ZSL and FSL.  

\begin{itemize}
\item \textbf{WordNet} is a large lexical database with several different relationships between words, such as synonym, hyponym, hypernym and meronym  \cite{miller1995wordnet}.
Its 3.0 version contains $155,287$ words, organized in $117,659$ synsets for a total of $206,941$ word-sense pairs.
WordNet can be directly accessed via online search and browse\footnote{\url{http://wordnetweb.princeton.edu/perl/webwn}}, and some python libraries such as NLTK. 
It is often used to build task-specific class hierarchies, especially for image classification, and has been the most widely used KG for augmenting both ZSL \cite{kampffmeyer2019rethinking,wang2018zero,liu2018combining,lee2018multi,wei2019residual,akata2015label,li2015zero,geng2020explainable,geng2021ontozsl,amador2021ontology,wang2021zero,chen2020zero} and FSL \cite{chen2020knowledge,tsai2017improving,peng2019few,jayathilaka2021ontology,monka2021learning,akata2015label}. 

\item \textbf{ConceptNet} is a freely-available Semantic Network with commonsense knowledge\footnote{\url{https://conceptnet.io/}} \cite{speer2017conceptnet}.
It stores a large number of entities which are either words or phrases.
The latest version ConceptNet 5 has around  $34$ million facts (relationships between entities) of $34$ relations including \textit{Synonym}, \textit{IsA}, \textit{RelatedTo}, \textit{HasContext}, \textit{HasA}, etc., and well supports $10$ core languages.
The \textit{IsA} relation represents hyponyms and hypernyms, from which class hierarchies are often used for augmenting ZSL \cite{nayak2020zero,zhang2019tgg,roy2020improving,chen2021zero,nguyen2021dozen,zhang2019integrating,chen2021zerotext} and FSL \cite{zhang2019tgg,yang2021empirical,zhang2019integrating}
It is mostly applied in CV tasks but has also been exploited in open information extraction (e.g., Nguyen et al. \cite{nguyen2021dozen}).

\item \textbf{Freebase} is a large-scale KG with relational facts, contributed by multiple sources such as Wikipedia, MusicBrainz (a music database), Notable Names Database (an online database of biographical details of famous people) and volunteers \cite{bollacker2008freebase}. 
Its official API has been shut down 2016, but it can still be accessed as a dump or via Google's Knowledge Graph API.
The dump on Google\footnote{\url{https://developers.google.com/freebase}} has around $1.9$ billion triples with tens of millions of entities, while $63$ million additional triples that have been deleted can also be downloaded. 
Freebase has been widely used for investigating KG techniques including KG augmented ZSL \cite{ma2016label,imrattanatrai2019identifying,amador2021ontology} and FSL \cite{zhang2021knowledge,ma2016label}.
Different from WordNet and ConceptNet, Freebase is mainly applied in open information extraction. 

\item \textbf{Wikidata} is a collaboratively edited KG that is increasing at a high speed. By November 2022, it has over $100$ million data items (entities).
Wikidata can be directly downloaded as a dump, or accessed via its official online SPARQL query service\footnote{\url{https://query.wikidata.org/}} and APIs. 
It is increasingly used in different applications, but its usage for augmenting ZSL and FSL had not attracted any attention until when two studies \cite{qu2020few,zhang2021knowledge} were proposed for augmenting few-shot relation extraction and another two studies \cite{geng2021ontozsl,li2020logic} were proposed for augmenting ZSL.

\item \textbf{DBpedia} is also a large-scale general purpose KG whose knowledge are mainly from Wikipedia, equipped with an ontological schema in OWL \cite{auer2007dbpedia}.
For the 2016-04 release, the English version has $6.0$ million things (entities) and $9.5$ billion RDF triples. DBpedia also has localized versions in 125 languages with much more entities.
DBpedia can be directly downloaded as a dump, or accessed via its online SPARQL query service\footnote{\url{https://dbpedia.org/sparql}} and lookup service/API\footnote{\url{https://lookup.dbpedia.org}} which can efficiently return a ranked list of entities for an input phrase. 
It has also been used to augment ZSL, often acting as a complement of relational facts and literals such as entity descriptions \cite{geng2020explainable,geng2021ontozsl,chen2021zero}. 
DBpedia's schema (ontology) can also provide hierarchical concepts and other schema knowledge for e.g., augmenting zero-shot KG completion with unseen entities \cite{amador2021ontology}.

\item \textbf{NELL} is a popular KG continuously extracted from the Web \cite{mitchell2018never}. 
According to its official statistics accessed in November 2022, it has accumulated $2.8$ million high confident beliefs (triples).
NELL can be browsed online\footnote{\url{http://rtw.ml.cmu.edu/rtw/kbbrowser}} or downloaded. 
We find two ZSL works and one FSL work that utilize NELL:
Wang et al. \cite{wang2018zero} use NELL for zero-shot classification for images from NEIL --- an image repository whose classes are aligned with NELL entities \cite{chen2013neil};
Geng et al. \cite{geng2021ontozsl} use its RDFS schema (ontology) for augmenting zero-shot KG completion with unseen relations;
Sui et al. \cite{sui2021knowledge} use its entity concepts for augmenting few-shot text classification.

\item \textbf{WebChild} a large collection of commonsense knowledge extracted from the Web as NELL \cite{tandon2014webchild}.
It contains triples that connect nouns with adjectives via fine-grained relations like hasShape, hasTaste, evokesEmotion, etc. 
Its 2.0 version has over $2$ million disambiguated concepts and activities (entities), connected by over $18$ million assertions (facts).
WebChild has now been rarely used in ZSL and FSL.
For augmenting (zero-shot) VQA, Chen et al. \cite{chen2021zero} and Wang et al. \cite{wang2017fvqa} use an auxiliary KG, whose facts are extracted from WebChild as well as ConceptNet and DBpedia.

\end{itemize}
}

\subsection{\rvc{KG Construction for Zero-shot and Few-shot Learning}}
Nowadays, there are many existing KGs which are constructed in different ways. 
\rvc{Most} high quality domain-specific ontologies such as the medical ontology SNOMED \cite{schulz2009snomed}
are often directly constructed by domain experts via collaboration, \rvc{while many aforementioned general purpose KGs} are constructed via crowdsourcing --- they are either extracted from existing crowdsourced resources like Wikipedia or directly contributed by volunteers.
To be more comprehensive, many KGs integrate different knowledge resources and databases; for example, ConceptNet \cite{speer2017conceptnet}, which was originally developed by crowdsourcing, further fused knowledge from DBpedia, Wiktionary, OpenCyc and so on.
In fact, solutions and technologies of Linked Open Data, Ontology Network and Ontology Alignment can all be used for constructing KGs via integration.
With the development of data mining, ML and other data analysis techniques, knowledge extraction from unstructured and semi-structured data such as the Web pages, tables and text have recently been widely investigated and used for KG construction; for example, NELL is continuously extracted from the Web \cite{mitchell2018never}, while Google's KG is extended with knowledge extracted from tables in Web pages \cite{cafarella2018ten}. 

For some specific \rvc{ZSL or FSL tasks}, there are exactly suitable KGs that can be directly applied.
For example, Huang et al. \cite{huang2018zero} directly \rvc{use} the event ontology named FrameNet \cite{baker2003framenet} for supporting their zero-shot event extraction method.
However, for the majority of the \rvc{ZSL and FSL} tasks, existing KGs usually cannot be directly applied due to their large sizes and irrelevant knowledge, and an (ad-hoc) KG should be extracted or constructed. 
In this part, we mainly review techniques of constructing KGs for augmenting \rvc{ZSL and FSL}.
We divide these techniques into three categories: sub-KG extraction from existing KGs, KG construction with task-specific data, and knowledge integration.

\subsubsection{Sub-KG Extraction}

Given a \rvc{ZSL or FSL} task, a straightforward solution is re-using an existing KG by extracting relevant knowledge.
\rvc{In the above part, we have already introduced those popular and general purpose KGs, and the ZSL and FSL studies where each KG is applied.
A ZSL or FSL method often extract a part of the KG by first matching the ML classes with KG entities, and then extracting the matched entities as well as some other related knowledge including neighbouring entities of the matched entities within k-hops, entities associated with the matched entities according to some specific relations, entities close to the matched entities in some embedding space, (hierarchical) concepts and other schema information of the entities, literals such as entity synonyms, descriptions and data properties, etc. 
Besides entities, some other KG elements such as relations and concepts can also be directly matched with related knowledge extracted. 
}

\rvc{
For some ZSL and FSL benchmarks, classes have already been matched with KG entities; for example, in the work \cite{wang2018zero}, a WordNet sub-graph with 30K nodes are extracted as a KG for an ImageNet subset that has 1K training classes, where all ImageNet classes are originally aligned with WordNet nodes.
For most other benchmarks, the matchings are built by simple name comparison or some knowledge retrieval services, sometimes with even human intervention. For example, Kampffmeyer et al. \cite{kampffmeyer2019rethinking} and Geng et al. \cite{geng2021ontozsl} manually match all the $50$ classes in an animal image classification benchmark named AwA2 with WordNet nodes; Nguyen et al. \cite{nguyen2021dozen}, who work on zero-shot entity extraction from text, first extract nouns and pronouns with a part-of-speech algorithm from all sentences in the dataset, and then search for their corresponding entities in ConceptNet and extract the matched entities and their adjacent ones.
}

\rvc{Besides, some other domain specific KGs have also been exploited for augmenting ZSL and FSL with a part of their knowledge.}
Zhang et al. \cite{zhang2021knowledge} \rvc{extract} concept-level relation knowledge from \textbf{UMLS} --- an ontology of medical concepts \cite{mccray2003upper}, for few-shot relation extraction in the medical domain.
Rios et al. \cite{rios2018few} \rvc{extract} class hierarchies and class descriptions from \textbf{ICD-9} diagnosis and procedure labels for zero-shot and few-shot medical text classification.
Luo et al. \cite{luo2020context} \rvc{extract} a sub-KG for object relationships from \textbf{Visual Genome} --- a knowledge base that stores connections between image visual concepts and language concepts \cite{krishna2017visual}, for augmenting zero-shot object recognition.
\cz{Zhou et al. \cite{zhou2021encoding} \rvc{train} their zero-shot question answering model with facts extracted from \textbf{WorldTree} (V2.0) \cite{xie2020worldtree} --- a knowledge base that contains explanations for multiple-choice science questions in the form of graph, covering both commonsense and scientific knowledge.
}

\subsubsection{Task-oriented KG Construction}
Instead of utilizing existing KGs, some ZSL and FSL studies build task-specific KGs.
The classes' textual information such as \rvc{class label} is the most frequently utilized information for mining inter-class relationships and further for constructing \rvc{the edges of a KG}.
Palatucci et al. \cite{palatucci2009zero} \rvc{connect} a word (which corresponds to a class in that task) to another according to their co-occurrence in a text corpus.
Lee et al. \cite{lee2018multi} \rvc{calculate} WUP similarity of class labels, and used this similarity to build KG edges for representing positive and negative inter-class relationships.
Wei et al. \cite{wei2019residual}, Ghosh et al. \cite{ghosh2020all} and Wang et al. \cite{wang2021zero} all \rvc{consider} calculating and adding edges to entities that are close to each other according to their labels' word embeddings.
Class attributes have also been exploited for mining inter-class relationships.
Zhang et al. \cite{zhang2019tgg} \rvc{build} a KG for the CUB benchmark which includes images of birds of fine-grained classes, by computing Hadamard product over the part-level class attributes.
Hu et al. \cite{hu2021graph} and Chen et al. \cite{chen2020zero} both directly \rvc{utilize} the co-occurrence of class attributes to build edges between KG entities.
Specially, Changpinyo et al. \cite{changpinyo2016synthesized} \rvc{consider} both class attributes and word embeddings to calculate weighted edges between entities.
Different from the above methods that use some auxiliary information for building KG edges, Zhao et al. \cite{zhao2020knowledge} and Geng et al. \cite{geng2020explainable,geng2021ontozsl} \rvc{model} the class attributes as additional KG entities and \rvc{connect} them to the entities \rvc{corresponding to} the classes; while Li et al. \cite{li2020transferrable,li2019large} \rvc{generate} new superclasses of the seen and unseen classes by clustering of the class names, so as to constructing class hierarchies for augmenting both ZSL and FSL. 

Domain knowledge, which is often in the form of heuristics and logic rules, has also be used to construct task-specific KGs.
Banerjee et al. \cite{banerjee2020self} \rvc{use} heuristics to create a synthetic KG with science facts from the QASC text corpus and commonsense facts from the Open Mind Commonsense knowledge (text) corpus, for addressing both zero-shot and few-shot question answering.
Chen et al. \cite{chen2020ontology} \rvc{add} existential restrictions (a kind of description logic that quantifies a class by associated properties) to some classes of an animal taxonomy extracted from WordNet, so as to build an OWL ontology for the animal image classification benchmark AwA2.

There are also some ZSL and FSL studies that extract structured knowledge from the task data (samples) for constructing KGs which are further fed back to learning for augmentation. 
When Ghosh et al. \cite{ghosh2020all} \rvc{construct} a KG for evaluating methods for few-shot action classification where some videos (samples) are given for each unseen class, they first extracted sample features for each class, then took the mean of these features as a KG entity, and finally calculated the cosine similarity between feature means for edges between entities.
Bosselut et al. \cite{bosselut2021dynamic} \rvc{generate} a temporary KG on demand for each prediction request of zeros-shot question answering, using its text context and a Transformer-based neural knowledge model named COMET \cite{bosselut2019comet}. 
Chen et al. \cite{chen2020zero} \rvc{add} a co-occurrence relation between two classes (food ingredients) by calculating their co-occurrence frequency in the training samples, besides the common class attributes and class hierarchies.

\subsubsection{Knowledge Integration}
Although some general purpose KGs contain a large quantity of knowledge and are being continuously extended, it is still common that the knowledge extracted from such a KG is incomplete or not fine-grained enough for a specific \rvc{ZSL or FSL} task.  
Therefore, some studies proposed to integrate knowledge extracted from different KGs or/and other resources for building a high quality task-specific KG.
For example, Chen et al. \cite{chen2021zero} \rvc{extract} RDF facts from three KGs --- ConceptNet, WebChild and DBpedia to generate a unified commonsense KG for augmenting zero-shot VQA;
Geng et al. \cite{geng2020explainable,geng2021ontozsl} \rvc{integrate} class hierarchies from WordNet, relational facts and literals from DBpedia, and knowledge transformed from class attributes for constructing KGs for zero-shot image classification;
Chen et al. \cite{chen2020zero} \rvc{consider and integrate} class hierarchies from WordNet, and class co-occurrence relations extracted from class attributes and samples for a KG for zero-shot ingredient recognition from food images.
Very recently, Geng et al. \cite{geng2023benchmarking} \rvc{conduct} a benchmarking study, where six KG equipped ZSL benchmarks \rvc{are} created for three different tasks and used for evaluating different methods under different auxiliary information settings.
The KG of each benchmark is based on the integration of multiple knowledge resources: those for zero-shot image classification \rvc{integrate} knowledge from WordNet, ConceptNet, class attributes, class names and so on, while those for zero-shot KG completion and relation extract \rvc{integrate} relation textual information, schema information from Wikidata or NELL, logic rules by human beings and so on.

As matching classes to KG entities for sub-KG extraction, the alignment of entities and relations in integrating different knowledge parts now is still mostly based on simple name matching or manual matching.
There is little attention to investigating automatic knowledge integration methods for \rvc{ZSL or FSL}, and the impact of the knowledge quality, such as the matching accuracy and the ratio of relevant or redundant knowledge, is often ignored.

\begin{figure}
\centering
\includegraphics[width=0.49\textwidth]{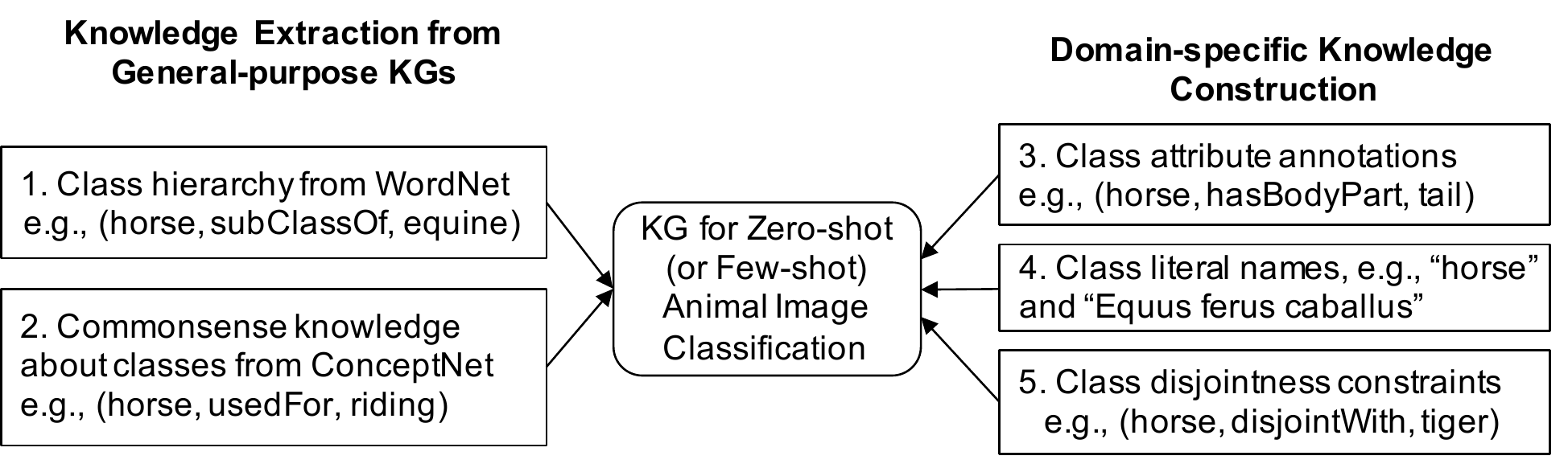}
\caption{\rvg{An example KG construction}. \label{fig:kg_case}}
\end{figure}

\rvg{\subsubsection{A Case of KG Construction}
Figure \ref{fig:kg_case} shows an example of constructing a KG with different semantics for augmenting a zero-shot (or few-shot) image classification task, which is from our previous benchmarking work \cite{geng2023benchmarking}.
First, the hierarchical relationship between classes and commonsense knowledge about classes are extracted from WordNet and ConceptNet, respectively, where classes are matched with KG entities.
For example, for an animal class \textit{horse}, its ancestors such as \textit{equine} are obtained from WordNet, and relational facts such as (\textit{horse, usedFor, riding}) are extracted from ConceptNet.
Next, some domain-specific knowledge such as attribute annotations of classes are represented as triples, where classes and attributes are represented as entities and connected via ad-hoc properties. For example, the attribute of \textit{tail} is related to \textit{horse} via property \textit{hasBodyPart}.
The literal names of classes are also represented using data properties.
More complex semantics such as the disjointness between classes are also represented using OWL.
For example, although \textit{horse} and \textit{tiger} have many shared attributes such as \textit{tail} and \textit{muscle}, they are still categorized as different species, and a disjointness constraint between them is added.
Finally, all kinds of class knowledge mentioned above are integrated into one KG to serve the ZSL (or FSL) task.
}

\section{KG-aware Zero-shot Learning}\label{sec:zsl}

According to the solutions for exploiting KGs, we divide the KG-aware ZSL methods into \rvc{four paradigms: \textit{Mapping-based}, \textit{Data Augmentation}, \textit{Propagation-based} and \textit{Class Feature}, as shown in Figure \ref{fig:categories}.
Table \ref{table:zsl} summarizes the paradigms and lists the papers of each category}.
We will next introduce the details of each paradigm.

\begin{table*}[t]
\footnotesize{
\centering
\renewcommand{\arraystretch}{1.8}
\begin{tabular}[t]{m{2.2cm}<{\centering}|m{5.6cm}<{\centering}|m{3.2cm}<{\centering}<{\centering}|m{4cm}<{\centering}}\hline
 \textbf{Paradigms} &\textbf{Summary} &\textbf{Categories} & \textbf{Papers}   \\ \hline
\multirow{3}{*}{Mapping-based} & \multirow{3}{*}{\makecell{These methods project the input and/or the \\ class into a common vector space where \\ a sample is close to its class w.r.t.  some \\ distance metric, and prediction can be \\ implemented  by searching the nearest class.}}  & Input Mapping & \cite{palatucci2009zero,li2015zero,ma2016label,liu2018combining,imrattanatrai2019identifying,chen2020ontology,li2020transferrable,li2020logic} \\ \cline{3-4}
& & Class Mapping & \cite{akata2013label,akata2015label,changpinyo2016synthesized,shah2019open,nayak2020zero}  \\  \cline{3-4}
& & Joint Mapping &  \cite{ma2016label,huang2018zero,hao2020inductive,roy2020improving,chen2021zero,rios2018few,chen2021zerotext}    \\  \hline
\multirow{2}{*}{Data Augmentation} & \multirow{2}{*}{\makecell{These methods generate samples or \\ sample features for the unseen classes, \\ utilizing KG auxiliary information.}} & Rule-based & \cite{rocktaschel2015injecting} \\ \cline{3-4}
 & & Generation Model-based & \cite{zhang2019tgg,qin2020generative,geng2021ontozsl} \\ \hline
\multirow{2}{*}{Propagation-based} & \multirow{2}{*}{\makecell{These methods propagate model parameters \\ or a sample's class beliefs from the  \\ seen classes to the unseen classes via a KG.}} & Model Parameter Propagation & \cite{wang2018zero,kampffmeyer2019rethinking,wei2019residual,geng2020explainable,chen2020zero,ghosh2020all,wang2021zero} \\ \cline{3-4}
& & Class Belief Propagation  &  \cite{lee2018multi,luo2020context,bosselut2021dynamic} \\ \hline
\multirow{2}{*}{Class Feature} & \multirow{2}{*}{\makecell{These methods encode the input and the class \\ into features often with their KG contexts \\ considered, fuse these features and feed them \\ directly into a prediction model.}} & Text Feature Fusion & \cite{zhao2017zero,shi2018open,logeswaran2019zero,yao2019kg,banerjee2020self,zhou2021encoding,niu2021open,wang2021kepler,amador2021ontology,wang2021inductive,zha2021inductive,wang2021structure,gong2021prompt}  \\ \cline{3-4}
& & Multi-modal Feature Fusion & \cite{amador2021ontology,nguyen2021dozen,zhang2019integrating,ristoski2021kg} \\ \hline

\end{tabular}
\vspace{0.1cm}
\caption{A summary of KG-aware ZSL paradigms.}\label{table:zsl}
}
\end{table*}

\subsection{Mapping-based Paradigm}
The mapping-based paradigm aims to build mapping functions towards the input (\rvc{$x$ or $\bm{x}$}) and/or the classes (\rvc{$y$ or $\bm{y}$}), so that their vector representations after mapping are in the same space and \rvc{are comparable
(i.e., an input is of a class if their vectors are close w.r.t. some metric like Cosine similarity and Euclidean distance)}.
We denote the mapping function for the input as $\mathcal{M}$ and the mapping function for the class as $\mathcal{M}'$.

\begin{figure*}
\centering
\includegraphics[width=0.9\textwidth]{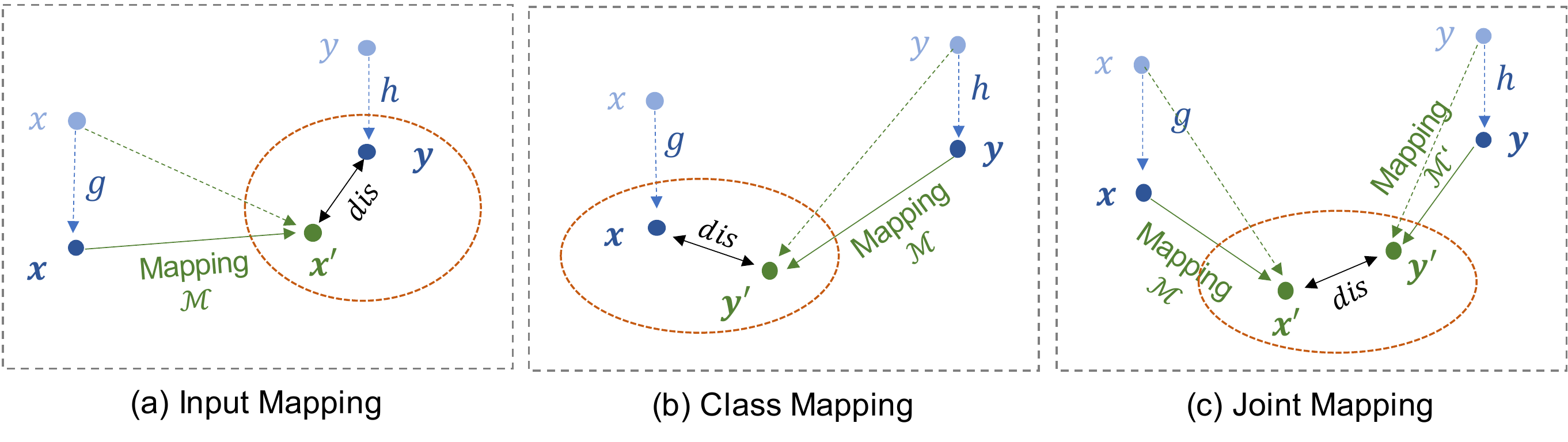}
\caption{Method categories and insights of the mapping-based paradigm. The dotted red circle denotes the vector space that the input and/or the class are mapped to. \label{fig:mapping}}
\end{figure*}

\rvc{$\mathcal{M}$ and $\mathcal{M}'$ can be both linear and non-linear transformation networks, often learned from the training data $\mathcal{D}_{tr}$.}
\rvc{Note that $\mathcal{M}$ and $\mathcal{M}'$ are different from the initial encoding functions $g$ and $h$. $g$ and $h$ are just to represent the symbolic input and class as vectors or to learn features and semantic embeddings, while $\mathcal{M}$ and $\mathcal{M}'$ mainly aim to map the input and the class into the same space, although sometimes they may also play the role of $g$ and $h$ at the same time.}
According to \rvc{the target(s) to map}, we divide the ZSL methods of the mapping-based paradigm into three \rvg{finer-grained} categories: \textit{Input Mapping}, \textit{Class Mapping} and \textit{Joint Mapping}. Figure \ref{fig:mapping} shows their insights.
\rvc{We will next introduce methods of each category, mainly from four dimensions --- input encoding, class encoding, mapping function(s) and comparison metric.}

\subsubsection{Input Mapping}\label{sec:input_mapping}
As shown in Figure \ref{fig:mapping} (a), \rvc{methods in this category only learn $\mathcal{M}$
to map the input into the space of the class initial encoding}.
\rvc{
A simple but typical method is proposed by Palatucci et al. \cite{palatucci2009zero} for neural activity classification.
In this task, the class is annotated by multiple attributes which are either calculated from classes' word similarity or manually created via crowdsourcing.
Each class is encoded by a multi-hot vector of its attributes\footnote{A multi-hot vector is to represent a set of discrete variables. Briefly, one slot corresponds to one variable; a slot is set to $1$ if its corresponding variable exists and to $0$ otherwise.}.
Given an input (neural activity signals) $x$, the mapping function $\mathcal{M}$, which is a multiple output linear regression model, predicts a multi-hot attribute encoding (vector) $\bm{y}'$. $\mathcal{M}$ is further attached by a 1-nearest neighbour classifier $\mathcal{L}$ which outputs the class as the one whose encoding is closest to $\bm{y}'$.
The whole model ($\mathcal{M}$ and $\mathcal{L}$) is jointly learned by minimizing some vector error-based loss on $\mathcal{D}_{tr}$.
}

\rvc{
Input mapping is widely used in zero-shot image classification, often with more complicated class encoding and mapping function than the method in \cite{palatucci2009zero}.
Li et al. \cite{li2015zero} map image features to the space of the class encoding, where is based on the word embeddings of the class itself and its super classes, and use the Cosine similarity for comparison.}
%
Chen et al. \cite{chen2020ontology} \rvc{adopt} a typical ZSL method named Semantic Autoencoder (SAE) \cite{kodirov2017semantic} \rvc{which uses a linear encoder as $\mathcal{M}$, and use pre-trained ontology embedding for the initial class encoding $\bm{y}$.
$\mathcal{M}$} is learned on $\mathcal{D}_{tr}$ by minimizing \gyx{a distance loss between $\bm{x}'$ and $\bm{y}$ and a reconstruction loss when $\bm{x}'$ is mapped back to $\bm{x}$.}
%
Li et al. \cite{li2020transferrable} \rvc{propose} to use a Long-Short-Term-Memory (LSTM) network to model the class hierarchy for \rvc{class encoding}, and map the image features learned by a CNN. 
%
Liu et al. \cite{liu2018combining} \rvc{use reinforcement learning and an ontology to associate rules with each class as visual characteristics}, and train one Support Vector Machine (SVM) for each rule as \rvc{the mapping function $\mathcal{M}$. $\mathcal{M}$ predicts associated rules of each input, and these predicted rules are compared with each candidate class.}
%

Input mapping has also been explored in \rvc{ZSL for knowledge extraction.}
\rvc{Ma et al. \cite{ma2016label} work on entity mention typing. They pre-train the initial class (entity type) encoding} using different KG embedding methods such as prototype-driven label embedding and hierarchical label embedding, and then proposed two mapping settings.
One setting is to directly \rvc{map the input (entity mention features) to the class encoding
by a linear transformation which is implemented by multiplying the input by a matrix of weights.}
Imrattanatrai et al. \cite{imrattanatrai2019identifying} \rvc{learn initial text representation of the input (relation mention) by word embedding and a Bidirectional LSTM network, and then use a linear transformation function to map the text representation into the space of relation encoding which is based on KG TransE embedding and ad-hoc relation feature extraction.}
%
Li et al. \cite{li2020logic} \rvc{map the input text representation into the class embedding by a simple linear transformation for zero-shot relation classification, where the class encoding combines word embedding, normal KG embedding and rule-guided KG embedding.}

\subsubsection{Class Mapping}
\rvc{The methods learn the mapping function $\mathcal{M}'$ to directly map the class into the space of the initial input encoding}, as shown in Figure \ref{fig:mapping} (b).
\rvc{It is not as widely investigated as input mapping.
In total, we gather three methods for zero-shot image classification \cite{akata2013label,akata2015label,changpinyo2016synthesized} and one method for zero-shot KG completion with unseen entities \cite{nayak2020zero}.}
Akata et al. \cite{akata2013label,akata2015label} 
learn \rvc{an embedding model} as the mapping function which \rvc{maps the class initial encoding into the space of the image features}.
They \rvc{use} class hierarchies as the auxiliary information \rvc{for initial class encoding, where each class is represented as a multi-hot vector of its ancestors.}
%
Changpinyo et al. \cite{changpinyo2016synthesized} first \rvc{generate} a weighted graph where the relatedness between classes are represented, then \rvc{introduce} phantom classes through which seen and unseen classes can be synthesized by convex combination, and finally \rvc{map} the vectors of phantom classes into the input.
%
Nayak et al. \cite{nayak2020zero} \rvc{propose} a novel transformer Graph Convolutional Network (GCN) \rvc{as $\mathcal{M}'$ which non-linearly aggregates a class's neighbours in the KG, and use a compatibility score as the metric for \rvc{comparing} the image CNN feature (input) and the class embedding.}
%
Shah et al. \cite{shah2019open} \rvc{predict} KG triples with unseen entities using their text descriptions. 
\rvc{The method first encodes the entity from the graph perspective by TransE or DistMult KG embedding (as initial class encoding $h$), and encodes the entity from the text perspective by word embedding and LSTM (as initial input encoding $g$), and then maps the class encoding to the space of the input encoding, where both linear and non-linear transformation functions such as Multi-Layer Perceptron (MLP) are explored.}

\subsubsection{Joint Mapping}
\rvc{As shown in Figure \ref{fig:mapping} (c), joint mapping learns both input mapping $\mathcal{M}$ and class mapping $\mathcal{M}'$ such that the input and the class are compared in one intermediate space.}
%
\rvc{It} is often adopted for zero-shot entity/relation extraction where features of both the input \rvc{(entity mention text)} and the class (entity/relation in a KG) are jointly \rvc{mapped}.
\rvc{Ma et al. \cite{ma2016label} multiply the initial input encoding and the initial class encoding by matrices (as $\mathcal{M}$ and $\mathcal{M}'$) which are learned by minimizing a weighted approximate-rank pairwise loss, for zero-shot entity extraction.
%
Huang et al. \cite{huang2018zero} \rvc{map the} features of event mentions and their structural contexts parsed from the text, and the event type encoding which is based on event ontology embedding, jointly into one vector space using one shared CNN.}

\rvc{Zero-shot text classification is very similar to zero-shot entity/relation extraction --- $\mathcal{M}$ and $\mathcal{M}'$ are applied to text input encoding and KG-based class encoding, respectively. 
Rios et al. \cite{rios2018few}  map the text features learned by a CNN, and the class encoding which is by initial word embedding and GCN-based class hierarchy embedding.
Chen et al. \cite{chen2021zerotext} linearly map the text encoding by BERT and the class encoding which is based on a word embedding model tailored by ConceptNet. 
}

\rvc{In zero-shot KG completion, the KG embedding (input) and the initial encoding of the unseen entity or relation (class) are jointly mapped.
Hao et al. \cite{hao2020inductive} \rvc{investigate} zero-shot KG completion with unseen entities. 
The input mapping $\mathcal{M}$ is a linear encoder over one-hot encoding of the KG entities, while the class mapping  $\mathcal{M}'$ is a MLP over the encoding of the entity's attributes.
}

%
\rvc{There are also some joint mapping methods in CV. 
Roy et al. \cite{roy2020improving} work on zero-shot image classification. 
They map both the initial class encoding learned by a GCN on commonsense knowledge and the image features learned by a CNN named ResNet101}, using a non-linear transformation named Relation Network.
This network first attaches a fully connected layer to the \rvc{class encoding}, then concatenates its output with the \rvc{image features}, and finally attaches two different fully connected layers. 
It is learned by minimizing a MSE loss. 
%
Chen et al. \cite{chen2021zero} work on \rvc{zero-shot VQA. They map the input (i.e., initial encoding of a pair of image and question) and the encoding of a KG entity (class) to a common space, where the matched KG entity is regarded as the answer.}

\subsection{Data Augmentation Paradigm}
A straightforward solution for addressing sample shortage is generating \rvc{new data with the guidance of some auxiliary knowledge.}
In ZSL, some methods generate samples or sample features for unseen classes and transform the problem into a standard supervised learning problem.
We regard these methods as Data Augmentation Paradigm.
\rvc{According to the generation method, we divide these methods} into two categories: \textit{Rule-based} and \textit{Generation Model-based}.

\subsubsection{Rule-based}
Background knowledge of a task could be explicitly represented by different kinds of rules (or \rvc{some} equivalent logic forms such as schema constraints and templates).
They enable deductive reasoning for hidden knowledge as new samples. 
\rvc{However, this solution has been rarely investigated for ZSL. 
In image classification and some other tasks where the sample input and their features are uninterpretable real value vectors, generating data by rules becomes unfeasible.
The only work we know is for zero-shot KG completion.
%
Rocktaschel et al. \cite{rocktaschel2015injecting} predict an unseen relation for two entity mentions extracted} from text.
They \rvc{propose} three methods to inject first-order rules, which act as commonsense knowledge, into a matrix \rvc{factorization-based KG completion model.
One of the methods} is logically inferring additional relational facts in advance before training the matrix factorization model.

\begin{figure}
\centering
\includegraphics[width=0.47\textwidth]{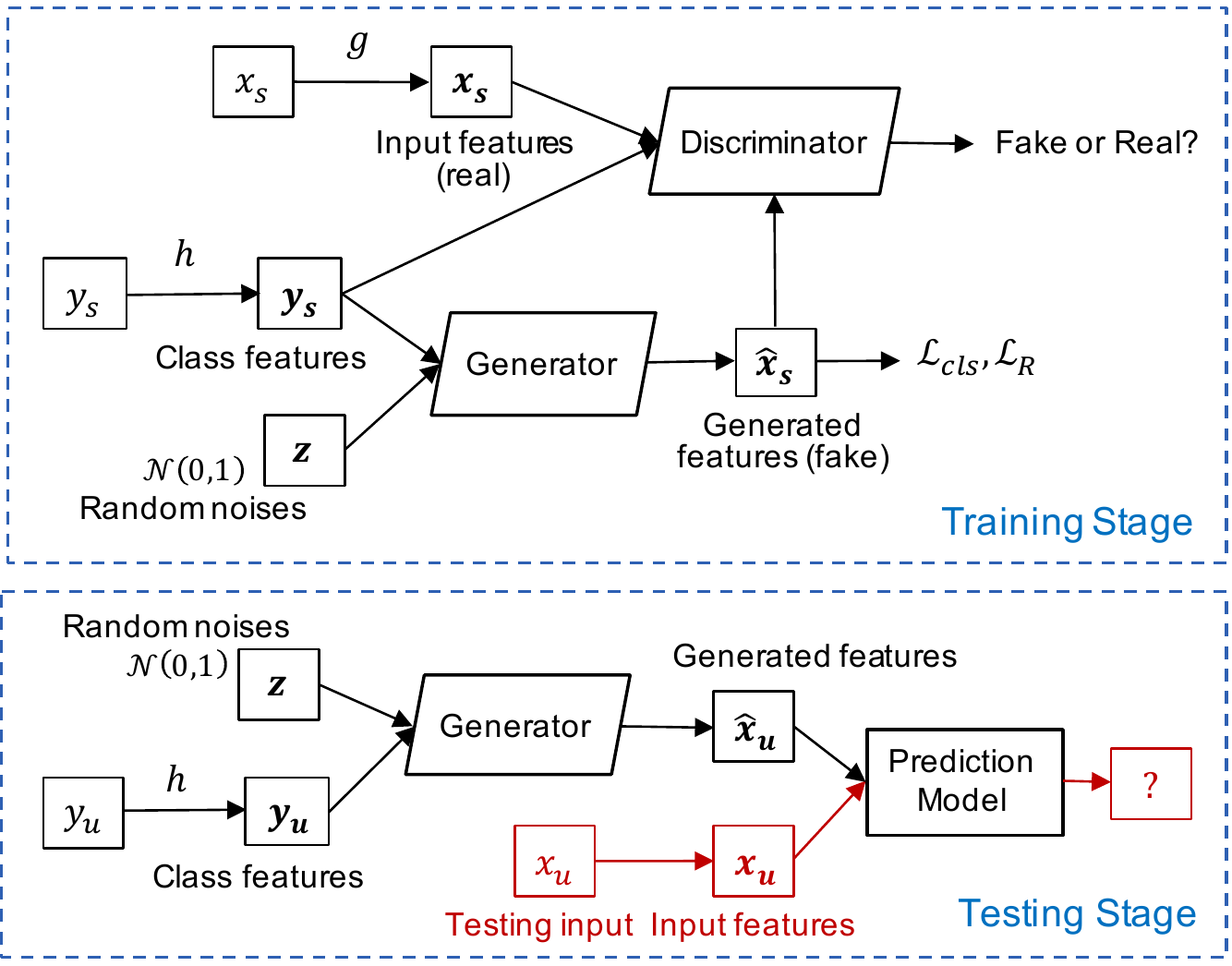}
\caption{\rvg{Overview of the GAN-based generation paradigm}. \label{fig:generation}}
\end{figure}

\subsubsection{Generation Model-based}
With the wide investigation of conditional generation, \rvc{generation} models such as Generative Adversarial Networks (GANs) \cite{goodfellow2014generative} and Variational Auto-encoder (VAE) \cite{kingma2013auto} have become popular tools for generating data for addressing ZSL especially for image classification \cite{geng2021ontozsl,huang2019generative,xian2018feature,zhu2018generative,chen2021knowledge,wang2019survey,zhang2019tgg}.
We \rvc{regard} these methods as Generation Model-based.
\rvg{In Figure~\ref{fig:generation}, we present a typical GAN-based scheme, including a training stage and a testing stage.
During training, given a seen class $y_s$, its encoding $\bm{y_s}$, which can be based on multi-hot attribute encoding, word embedding and KG embedding, is fed into the generator of the GAN, together with a random noise vector $\bm{z}$ sampled from Normal distribution $\mathcal{N}(0,1)$, to generate a set of sample features $\hat{\bm{x}}_s$ for $y_s$. Note the number of generated samples is a hyperparameter that can be tuned.
Correspondingly, a set of real features (encoding) $\bm{x}_s$ are extracted from the input samples $x_s$ of $y_s$ to supervise the training of the generator via an adversarial discriminator which distinguish $\bm{x}_s$ and $\hat{\bm{x}}_s$.
Both the generator and the discriminator are conditioned on the class embeddings. Neural networks composed of several fully connected layers are often used as their model.
Some additional loss terms such as classification ($\mathcal{L}_{cls}$) and regularization ($\mathcal{L}_R$) are usually also applied to encourage the model to generate more plausible samples.
During testing, the trained GAN can synthesize samples for an arbitrary unseen class $y_u$ with random noises and its encoding $\bm{y_u}$.
With the synthesized data $\hat{\bm{x}}_u$, we can learn classifiers for the unseen classes and use them to predict testing samples, as normal supervised learning. We can also directly compare the features of each input testing sample with the synthesized sample features of each unseen class to determine the output class label.}

\rvc{Conditional generation models were not widely applied in ZSL until around 2018. We find three KG-aware methods.} 
\rvg{Geng et al. \cite{geng2021ontozsl} is the first work that propose to generate the image features using a KG which models the semantic relationships between seen classes and unseen classes for zero-shot image classification. The class encoding based on this KG's text-aware embeddings leads to higher-quality image features than previous class encodings based on attributes and word embeddings.
GAN is used as the generation model.
The follow-up work by Geng et al. \cite{geng2022disentangled} further considers a new disentangled KG embedding method for encoding class semantics from multiple aspects, leading to better zero-shot image classification performance.
Qin et al. \cite{qin2020generative} work on a zero-shot KG completion problem, where the testing triples involve new relations that have never appeared in training. They propose to first use GAN to generate sample features (i.e., KG embeddings) of the unseen relations conditioned on their textual description embeddings (i.e., class encodings), and then calculate scores of the testing triples by comparing the generated relation embeddings with the existing embeddings of the entities.
%
%
Note the two works raised by Geng et al. \cite{geng2021ontozsl,geng2022disentangled} also deal with the unseen relations in zero-shot KG completion besides the zero-shot image classification.
They propose to synthesize the relation embeddings conditioned on the embeddings of an ontological schema which represents the semantics of the KG relations, such as relation hierarchy, and relations' domain and range constraints.}
Zhang et al. \cite{zhang2019tgg} \rvc{work} on zero-shot image classification by \rvc{generating few-shot samples.
They use} a generation module to generates an instance-level graph, where dummy features (instances) are synthesized for those unseen classes by GAN.
They finally address \rvc{the problem} over the instance-level graph by a propagation module and a meta learning strategy.
\rvg{Across the literature, we find that there is no current work that combines the VAE-based generation models with KG, leaving a great research space to explore.}

\subsection{Propagation-based Paradigm}

\rvc{Graph-based information propagation is a straightforward solution to utilize KG auxiliary information for ZSL. We regard these KG-aware ZSL methods as Propagation-based Paradigm.}
These methods \rvc{first} align seen and unseen classes with KG entities \rvc{and build a graph with node features from the auxiliary KG.
Then they use a graph propagation model to either approximate model parameters or predict class beliefs (scores) of the unseen classes, where this propagation model is usually trained from  seen classes whose outputs are given based on models built from $\mathcal{D}_{tr}$.}
\rvc{Accordingly, we divide the methods into \textit{Model Parameter Propagation}  and \textit{Class Belief Propagation}}.
\rvc{We will next introduce the general idea of each category, and its contributes mainly from two dimensions --- the propagation graph and the propagation model.}

\begin{figure*}
\centering
\includegraphics[width=0.8\textwidth]{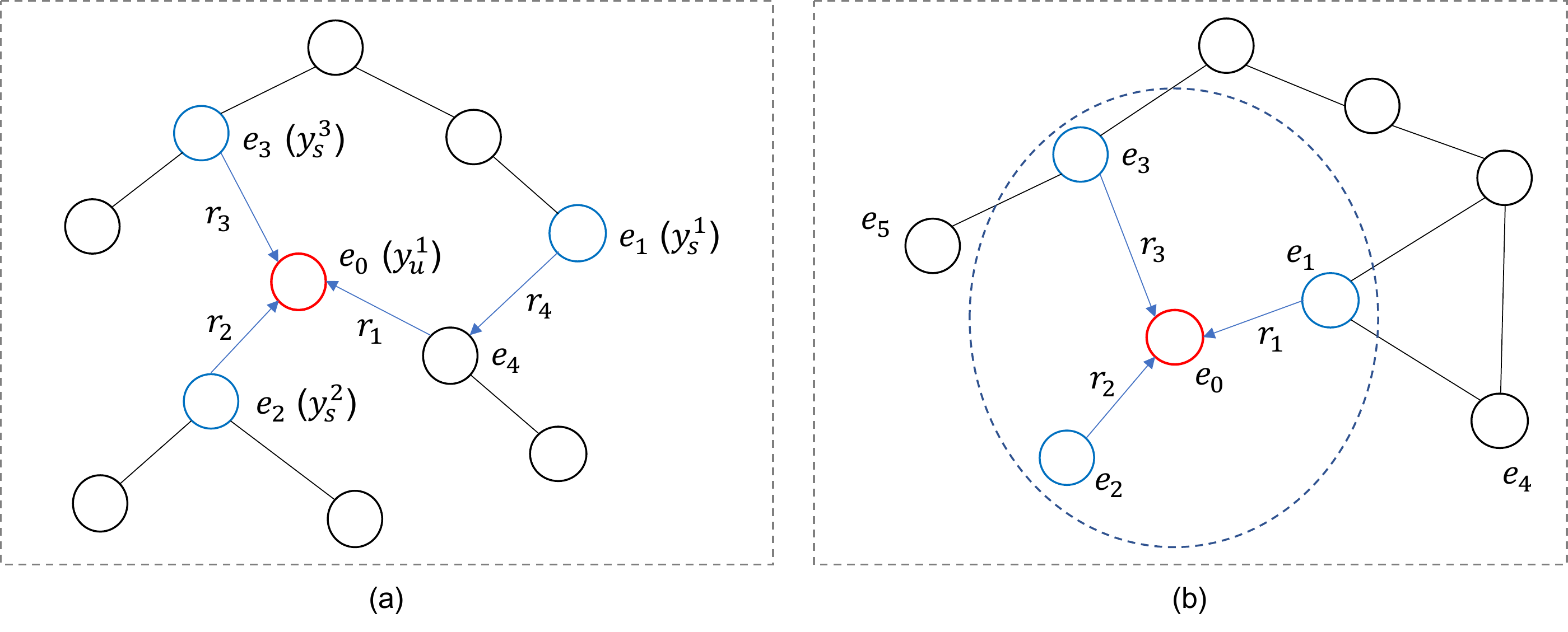}
\caption{(a) \rvc{Graph information propagation from entities of seen classes (e.g., $e_1$, $e_2$ and $e_3$) to entities of unseen classes (e.g., $e_0$) for approximating model parameters (or predicting class beliefs).} (b) \rvc{Aggregating embeddings of of 1-hop neighbouring seen entities (in blue) to get the embedding of an unseen entity (in red) for few-shot KG completion.}
 \label{fig:propagation}}
\end{figure*}

\subsubsection{Model Parameter Propagation}
\rvc{For each seen class $y_s$ in $\mathcal{Y}_s$, a one-vs-rest binary classifier is trained with $\mathcal{D}_{tr}$. Such a classifier is usually composed of a pre-trained input encoding $g$ (e.g., a CNN for image feature learning) and linear or non-linear classification layer(s). Parameters of the classification layer(s) are denoted as $p(y_s)$.
Considering a simple but general case in Figure \ref{fig:propagation} (a), the seen classes $y_s^1$, $y_s^2$ and $y_s^3$ are aligned with three graph entities $e_1$, $e_2$ and $e_3$, respectively. The parameters $p(y_s^1)$, $p(y_s^2)$ and $p(y_s^3)$ are assigned to $e_1$, $e_2$ and $e_3$ as their outputs, and the output parameters of $e_0$ which is aligned with an unseen class $y_u^1$ ($y_u^1 \in \mathcal{Y}_u$), i.e., $p(y_u^1)$, are approximated according to a graph propagation model. With $p(y_u^1)$ and the pre-trained input encoding $g$, samples of $y_u^1$ can be predicted.
The graph propagation model is usually trained by minimizing errors when approximating parameters of classifiers of $\mathcal{Y}_s$.}

%
In image classification, \rvc{usually one classifier, which is composed of a linear layer and pre-trained CNN image features, is trained for each seen class via $\mathcal{D}_{tr}$, and the linear layer parameters of the seen classes are propagated.}
\rvc{Wang et al. \cite{wang2018zero} adopt a CNN named ResNet-50 for image feature learning. The method aligns image classes with WordNet \cite{miller1995wordnet} entities, and uses a GCN to propagate feature combination weights on a sub-graph of WordNet.}
%
Wei et al. \cite{wei2019residual} \rvc{aim at} the same problem as \cite{wang2018zero}, but \rvc{use} a Residual Graph Convolutional Network (ResGCN), which \rvg{builds} residual connections between hidden layers, \rvc{for propagation} so as to alleviate over-smoothing and over-fitting \rvg{when stacking multiple GCN layers}.
Ghosh et al. \cite{ghosh2020all} \rvc{uses a 6-layer GCN for propagation on a KG for zero-shot action recognition (video classification)}.
%
Wang et al. \cite{wang2021zero} \rvc{construct} two single-relation KGs --- one \rvc{by} the class hierarchy from WordNet and the other \rvc{by} the class correlation mined from word embeddings for zero-shot image classification.
They \rvc{use} two weight-shared GCNs to propagate on the two KGs to \rvc{predict} classifier parameters for the unseen classes. 
%
\rvc{When propagating on a sub-graph of WordNet for zero-shot image classificaton, Geng et al. \cite{geng2020explainable} attach an attention layer after GCN layers to calculate the importance weights of different seen classes to an unseen class, which also provides a way for explanation \rvg{on feature transferability}.}
\rvc{Chen et al. \cite{chen2020zero} develop a propagation-based method for estimating} the parameters of multi-label classifiers for zero-shot ingredient recognition from food images.
Since the KG, which is composed of knowledge of ingredient hierarchy, ingredient attributes and ingredient co-occurrence, has multiple different relations, an attentive multi-relational GCN is adopted, where different relations have different contributions in parameter propagation.

\rvg{When propagating knowledge to distant nodes, all the above methods prefer to stack mutiple GCN layers. In contrast, Kampffmeyer et al. \cite{kampffmeyer2019rethinking} propose to only use two GCN layers and extend to directly connect an entity to its ancestors and descendants, where an attention mechanism is used to weight the contributions of different neighbouring entities according to their distances to the target entity.}
\rvc{Under the same task and evaluation setting as in \cite{wang2018zero},}
\rvg{the two-layer network and these weighted additional connections often achieve better performance.}

\subsubsection{Class Belief Propagation}
\rvc{These methods are often for multi-label classification where one sample is annotated by multiple related classes.
Without losing the generality, we assume a sample should be annotated by three seen classes ($y_s^1$, $y_s^2$ and $y_s^3$) and one unseen class ($y_u^1$), and these classes are matched to graph nodes, as shown in Figure \ref{fig:propagation} (a).
The beliefs (scores) of $y_s^1$, $y_s^2$ and $y_s^3$ are predicted by their corresponding binary classifiers trained with $\mathcal{D}_{tr}$, while the score of $y_u^1$ is predicted by a graph propagation model trained with outputs of nodes of the seen classes.
}

One typical \rvc{work} is the zero-shot multi-label image classification \rvc{method} by Lee et al. \cite{lee2018multi}, where multiple \rvc{objects are to be recognized from an input image.}
\rvc{It uses a gated recurrent update mechanism for iterative belief propagation on the auxiliary KG, where the propagation is directional from seen classes to unseen classes, and a standard fully-connected neural network to output a final belief for the entity of each unseen class.}
Note that the initial status of an entity is determined \rvc{by the features of the corresponding class's samples and word embedding.}
Luo et al. \cite{luo2020context} \rvc{work on} recognizing multiple interactive objects in an image where some objects are unseen in training.
They \rvc{use} Conditional Random Field to infer the unseen objects using the recognized seen objects in the image and a KG with prior knowledge about the relationships between objects.
Bosselut et al. \cite{bosselut2021dynamic} \rvc{focus} on zero-shot question answering. 
They propose to construct a context-relevant commonsense KG from deep pre-trained language models, where the question acts as a root entity and the \rvc{answers} act as leaf entities, and then they infer over the graph by aggregating paths to find the right answer.
\rvc{Although this method predicts only one answer (class) for each question (sample)}, but associates one question with multiple candidate answers \rvc{for graph information propagation}.

\subsection{Class Feature Paradigm}\label{sec:zsl_class_feature}
\begin{figure}
\centering
\includegraphics[width=0.48\textwidth]{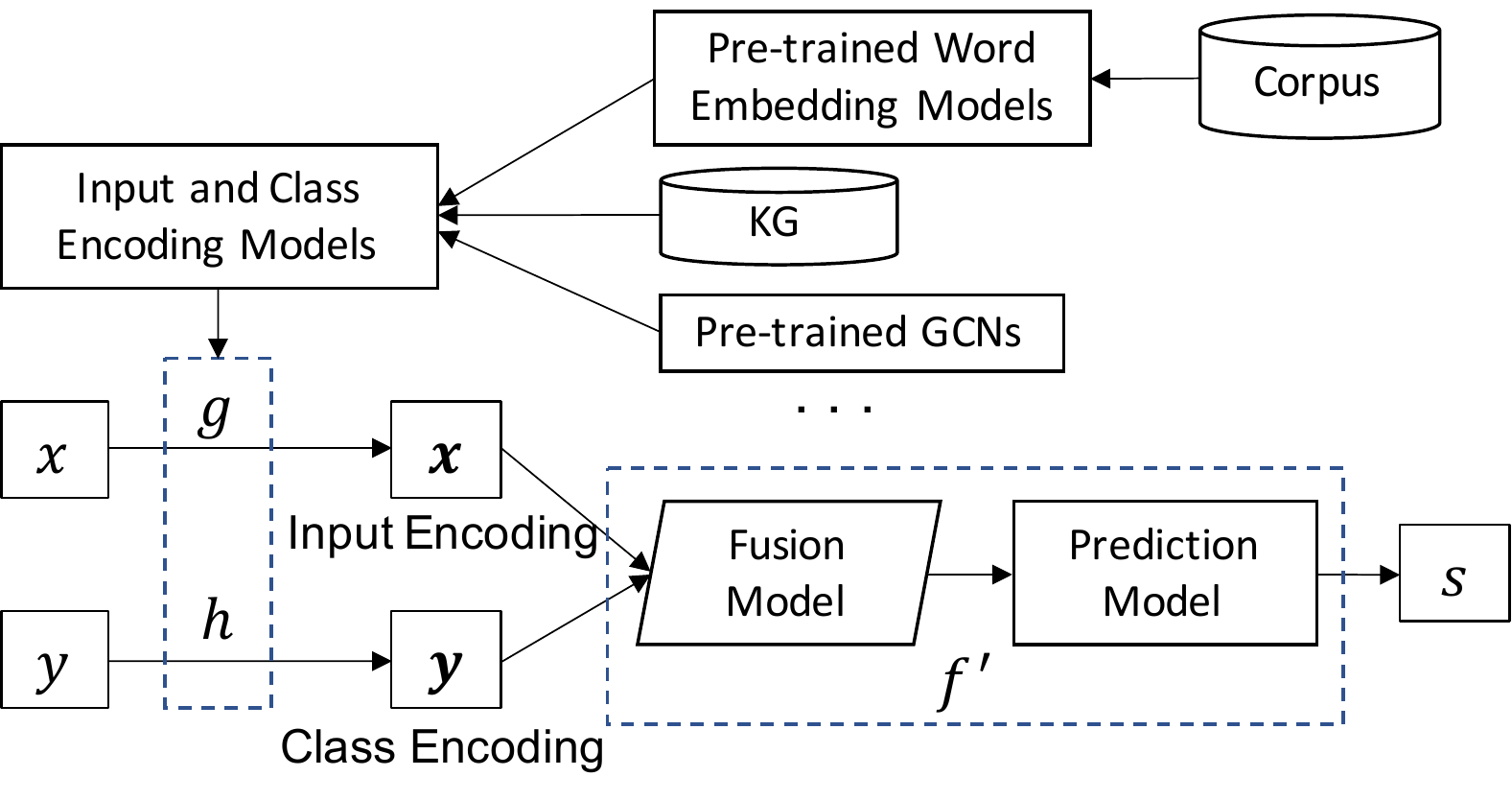}
\caption{\rvc{Overview of the class feature paradigm}. \label{fig:class_feature}}
\end{figure}

\rvc{As shown in Figure \ref{fig:class_feature}, class feature paradigm uses a transformed scoring function $f': (g(x), h(y)) \rightarrow s$ to calculate a matching score $s$ between an input $x$ and a class $y$.
The KG auxiliary information is usually utilized by the class encoding $h$.
$f'$ is usually composed of a fusion model, which combines $g(x)$ and $h(y)$, and a prediction model. $f'$ can be trained with $\mathcal{D}_{tr}$. $h$ and $g$ can be separately pre-trained or trained jointly with $f'$.}


This \rvc{paradigm} actually transforms the ZSL problem into a classic \textit{domain adaption} problem: \rvc{considering the new input of $f'$, the training data ($y \in \mathcal{Y}_{s}$) have a different distribution as the testing data \rvc{($y \in \mathcal{Y}_{u}$ or $y \in \mathcal{Y}_{u} \cup \mathcal{Y}_s$)}}.
According to the types of \rvc{$g(x)$ and $h(y)$}, we further classify \rvc{these} KG-aware ZSL methods into two categories: \textit{Text Feature Fusion} and \textit{Multi-modal Feature Fusion}.
\rvc{Next we will introduce the works of each category mainly from three dimensions --- the class encoding, the input encoding, and the fusion model.}

\subsubsection{Text Feature Fusion}
\rvc{These methods usually aim at ZSL tasks within a KG context where the auxiliary information is some kind of text.
One typical setting} is KG completion with unseen entities \rvc{where entities are} described by name phrases and/or textual descriptions.
\rvc{In this case, the class, i.e., an entity (or a relation) in a triple to predict, and the input, i.e., the remaining two elements of the triple, are both represented as in a textual form and encoded by text embedding.}

Zhao et al. \cite{zhao2017zero} \rvc{adopt} the TF-IDF algorithm to combine the embeddings of words to \rvc{encode} each entity with its text description.
For \rvc{a triple, the scoring function $f'$ uses the encodings of its two entities to calculate a triple score, where the interactions between any two elements of the triple are modeled.}
Shi et al. \cite{shi2018open} \rvc{propose} a zero-shot KG completion method named ConMask for dealing with unseen entities using their names and text descriptions.
\rvc{They feed the text encodings of the entities and the relation of a triple into a CNN-based fusion model.}
Niu et al. \cite{niu2021open} \rvc{follow} the general direction of \cite{zhao2017zero} and \cite{shi2018open}, but \rvc{work} out a new multiple attention-based method with a Bidirectional LSTM network and an attention layer for modeling and utilizing the interaction between the head entity description, head entity name, the relation name, and the tail entity description.
Amador et al. \cite{amador2021ontology} \rvc{work} on triple classification with unseen entities. 
The ontological information such as the entity's hierarchical classes are utilized \rvc{with} their word embeddings.
\rvc{For each entity, its hierarchical classes' word embeddings are combined with its own word embedding} by concatenation, averaging or weighted averaging.
Wang et al. \cite{wang2021inductive} \rvc{propose a KG completion method InductiveE which can deal with unseen entities using} entity textual descriptions.
It \rvc{encodes an entity by concatenating its text embeddings} by the fastText word embedding model \cite{joulin2016fasttext} and the pre-trained BERT \cite{devlin2019bert}. \rvc{For each triple, it feeds the entities' encodings} into a model composed of an encoder --- a gated-relational GCN and a decoder --- a simplified version of ConvE \cite{dettmers2018convolutional} to predict \rvc{a} score. 

Recently, due to the wide investigation of pre-trained language models, some methods that fine-tune these models for utilizing textual information for addressing zero-shot KG completion have been proposed.
\rvc{Different from \cite{wang2021inductive} where BERT is used for initial but fixed entity encoding, the entity and relation encoding in the following methods are trained jointly with $f'$ as the pre-trained language model} BERT is fine-tuned.
\rvc{Yao et al. \cite{yao2019kg} propose a KG triple prediction method called KG-BERT. It transforms each triple into a text sentence with the name information of its head entity, predicate and tail entity, and then makes triple prediction as a downstream text classification task, where BERT is fine-tuned with triples for training.}
Zha et al. \cite{zha2021inductive} also \rvc{propose to predict triples as a downstream text classification task of BERT. They fine-tune BERT using not only single triples but also possible paths that connect two entities (reasoning is conducted explicitly to discover usch paths).
Wang et al. \cite{wang2021structure} attempt to address two cons of KG-BERT: the combinatorial explosion in triple inference and the} failure to utilize structured knowledge.
They \rvc{propose} a structure-aware encoder to represent a triple's text with different combinations and interactions between its entities and relations.
Wang et al. \cite{wang2021kepler} \rvc{propose} a joint text and entity embedding method named KEPLER which is also able to predict KG triples with unseen entities and relations.
It utilizes \rvc{the text of the entities and the relations to fine-tune the BERT model via a masked language modeling loss.}

Besides KG completion, we also find \rvc{two KG-aware zero-shot question answering methods and on KG-aware zero-shot knowledge extraction method that belong to text feature fusion.}
Banerjee et al.  \cite{banerjee2020self} \rvc{perform} question answering via triple learning where the context, question and answer are modeled as a triple, and \rvc{one of them is predicted given the other two.}
In implementation, a transformer-based model that generates the answer given the text features of the context and question is learned by span masked language modeling, using triples extracted from text.
\rvc{Zhou et al. \cite{zhou2021encoding} also model} question answering as triple prediction with all the text features fused, and learn the prediction model by alternatively masking the subjects and the objects of the training triples which are from a corpus named WorldTree.
Gong et al. \cite{gong2021prompt} \rvc{fine-tune} a BERT model for zero-shot relation extraction, where prompts \rvc{are} constructed as the input using the relation's corresponding knowledge in ConceptNet.

\subsubsection{Multi-modal Feature Fusion}
\rvc{In these methods, the input encoding and the class encoding are of different types. Due to the heterogeneity of the inputs, the mechanism of $f'$, especially the fusion model, will differ from that of the text feature fusion category.
%
Nguyen et al. \cite{nguyen2021dozen} work on cross-domain entity recognition from the text, where the testing entities are from a different domain and unseen,
The input is a sequence encoded as token features by a pre-trained BERT, while the class is an entity encoded as graph features learned by a Recurrent GNN over an ontology.}
These two different kinds of features are fed into an integration network \rvc{for fusion.}
Ristoski et al. \cite{ristoski2021kg} \rvc{work on zero-shot entity extraction. They fuse the features of the entity mention and entity description (input), with the entity's graph vector (output) which encodes the entity's KG semantics such as types.}
%
Zhang et al. \cite{zhang2019integrating} \rvc{work} on zero-shot text classification. They \rvc{fuse the input text encoding and the ConceptNet-based class encoding, and feed them into a CNN classifier. The class encoding encodes the associated entity of the class, its ancestors and its description entities, using multi-hot encoding.}
%

\section{KG-aware Few-shot Learning}\label{sec:fsl}

\rvc{Many KG-aware FSL methods also follow the four paradigms of KG-aware ZSL}: \textit{Mapping-based}, \textit{Data Augmentation}, \textit{Propagation-based} and \textit{Class Feature}.
However, some other KG-aware FSL methods, 
belong to none of the above.
Instead, we regard those that focus on utilizing the few-shot samples by accelerating the adaption in training with meta learning algorithms, as a new paradigm named \textit{Optimization-based},
and regard those that directly transfer models (such as rules) built according to data of seen classes as another new paradigm named \textit{Transfer-based}.
\rvc{Figure \ref{fig:categories} presents these paradigms and their method categories while Table \ref{table:fsl} summarizes the paradigms and lists the papers of each category}.
We will next introduce the details of each paradigm.

\begin{table*}[t]
\footnotesize{
\centering
\renewcommand{\arraystretch}{1.8}
\begin{tabular}[t]{m{2.2cm}<{\centering}|m{5.8cm}<{\centering}|m{3.2cm}<{\centering}<{\centering}|m{4cm}<{\centering}}\hline
 \textbf{Paradigm} &\textbf{Summary} &\textbf{Categories} & \textbf{Papers}   \\ \hline
\multirow{3}{*}{Mapping-based} & \multirow{3}{*}{\thead{These methods project the input and/or the \\ class into a common vector space where \\ a sample is close to its class w.r.t.  some \\ distance metric, and prediction can be \\ implemented  by searching the nearest class. \\ ZSL methods can often be extended for FSL.}}  & Input Mapping & \cite{jayathilaka2021ontology,ma2016label,monka2021learning} \\ \cline{3-4}
& & Class Mapping & \cite{li2020transferrable} \\  \cline{3-4}
& & Joint Mapping &    \cite{akata2013label,ma2016label,akata2015label,xiong2018one,li2019large,zhao2020knowledge,zhang2020fewa,sui2021knowledge,zhang2021knowledge,rios2018few}  \\  \hline
\multirow{2}{*}{Data Augmentation} &  \multirow{2}{*}{\thead{These methods generate additional samples or \\ sample features for the unseen classes, \\ utilizing KG auxiliary information.}} & Rule-based & \cite{tsai2017improving}  \\ \cline{3-4}
& &Generation Model-based  & \cite{wang2019tackling,zhang2020relation} \\ \hline
\multirow{2}{*}{Propagation-based} & \multirow{2}{*}{\thead{These methods propagate model parameters,\\ or class embeddings (or a sample's class beliefs) \\ from the  seen classes to the unseen classes \\ via a KG.}} & \thead{Model Parameter \\ Propagation} & \cite{peng2019few,chen2020knowledge} \\ \cline{3-4}
& & Embedding Propagation  & \cite{hamaguchi2017knowledge,wang2019logic,albooyeh2020out,bhowmik2020explainable,zhao2020attention,dai2020inductively,ali2021improving} \\ \cline{1-4}
\multirow{2}{*}{Class Feature} & \multirow{2}{*}{\thead{These methods encode the input and the class \\ into features often with their KG contexts \\ considered, fuse these features and feed them \\ directly into a prediction model.}} & Text Feature Fusion & \cite{banerjee2020self}  \\ \cline{3-4}
& & \thead{Multi-modal Feature \\ Fusion} & \cite{zhang2019long,yang2021empirical,marino2021krisp}\\ \hline
\multirow{2}{*}{Optimization-based} & \multirow{2}{*}{\thead{These methods adopt meta learning algorithms \\ to optimize the training that relies on \\ the few-shot samples.}} & KG-specific Optimization & \cite{wang2019meta,chen2019meta,baek2020learning} \\ \cline{3-4}
& & KG-agnostic Optimization &\cite{lv2019adapting,zhang2020fewb,qu2020few,zhang2020fewa,zhang2019tgg,zhang2021knowledge} \\ \hline
\multirow{2}{*}{Transfer-based} & \multirow{2}{*}{\makecell{These methods directly apply models of seen \\ classes to unseen classes, often with the \\ few-shot samples utilized in prediction.}} & Neural Network Transfer & \cite{teru2020inductive,liu2021indigo,chen2021topology}   \\ \cline{3-4}
 & & Rule Transfer & \cite{sadeghian2019drum,mihalkova2007mapping,mihalkova2009transfer,davis2009deep,van2015todtler}    \\ \hline
\end{tabular}
\vspace{0.1cm}
\caption{A summary of KG-aware FSL paradigms.}\label{table:fsl}
}
\end{table*}

\subsection{Mapping-based Paradigm}

The general idea of the mapping-based paradigm of FSL is very close to that of ZSL, \rvc{which is to train an input mapping model $\mathcal{M}$, a class mapping model $\mathcal{M}'$ or both $\mathcal{M}$ and $\mathcal{M}'$ as shown in Figure \ref{fig:mapping}}. 
In contrast to ZSL, FSL has a small number of labeled samples associated with each unseen class \rvc{($\mathcal{D}_{few}$)}.
They usually can play an important role and are fully utilized.
\rvc{For example, they are sometimes used as another kind of auxiliary information of the classes.}
\rvc{As ZSL, we further categorize the mapping-based KG-aware FSL methods into \textit{Input Mapping}, \textit{Class Mapping} and \textit{Joint Mapping}.
Next we will introduce the methods of each category, from the dimensions of input encoding, class encoding, mapping function(s), comparison metric and few-shot sample utilization.}

\subsubsection{Input Mapping}\label{sec:fsl_input_mapping}
ZSL methods of input mapping can often be directly extended to FSL by augmenting the learning of the mapping model \rvc{$\mathcal{M}$} with the few-shot samples.
Ma et al. \cite{ma2016label} \rvc{support zero-shot and few-shot entity mention typing at the same time. They get the initial class (entity type) encoding via KG embedding and then directly map the input encoding (entity mention features) to the space of the class encoding. In the few-shot setting, they train the mapping mode $\mathcal{M}$, i.e., a linear transformation function with both training samples $\mathcal{D}_{tr}$ and few-shot samples $\mathcal{D}_{few}$.}
Jayathilaka et al. \cite{jayathilaka2021ontology} also \rvc{learn a mapping model $\mathcal{M}$ using both $\mathcal{D}_{tr}$ and $\mathcal{D}_{few}$.
They represent logical relationships such as class disjointness and class subsumption by an OWL ontology and embed it by a logic embedding algorithm named EL Embedding \cite{kulmanov2019embeddings} for initial class encoding.}
Monka et al. \cite{monka2021learning} \rvc{investigate KG-augmented few-shot image classification.
They use a KG curated by experts for modeling the relationship between classes, embed the KG by a variant of GCN for class encoding, and adopt a contrastive loss to train a MLP as $\mathcal{M}$ for mapping the image features.
}

\subsubsection{Class Mapping}
\rvc{The learning of the class mapping model $\mathcal{M}'$ can also be directly extended using both $\mathcal{D}_{tr}$ and $\mathcal{D}_{few}$.}
However, such extension has been rarely investigated.
\gyx{\rvc{The only KG-aware FSL method that belongs to class mapping} is by Li et al. \cite{li2020transferrable}. \rvc{It maps} the embeddings of hierarchical classes \rvc{(class encoding) into the space of image CNN features (input encoding) for both zero-shot and few-shot image classification.}
The mapping function learning does not use \rvc{$\mathcal{D}_{few}$},
but in prediction,}
the average of the CNN features of the few-shot samples as well as the \rvc{mapped class vector are both used and compared with the input of a testing sample}.

\subsubsection{Joint Mapping}\label{sec:fsw_joint_mapping}
\rvc{Some KG-aware FSL methods of joint mapping are also simple extensions of their ZSL counterparts}.
%
Akata et al. \cite{akata2013label,akata2015label} \rvc{jointly map the WordNet-based class encoding and the image encoding into one common space for few-shot image classification, where the mapping models $\mathcal{M}$ and $\mathcal{M}'$ are trained with an additional loss on few-shot samples $\mathcal{D}_{few}$.
Similarly, Ma et al. \cite{ma2016label} utilize $\mathcal{D}_{few}$ to augment the training of the mapping models for few-shot entity typing.}
\rvc{Note they consider not only input mapping, but also joint mapping.
%
Rios et al. \cite{rios2018few} also augment the joint training of $\mathcal{M}$ and $\mathcal{M}'$ with $\mathcal{D}_{few}$, for KG-aware few-shot text classification, where the input text encoding is based on CNN while the class encoding uses GCN.
}

\rvc{Some other KG-aware FSL methods of joint mapping are specifically developed for utilizing the few-shot samples $\mathcal{D}_{few}$.
They regard $\mathcal{D}_{few}$ as a kind of auxiliary information, and map them and the testing samples into one common vector space.}
Li et al. \cite{li2019large} jointly \rvc{learn a mapping of the image CNN features (input encoding) and a mapping of the class encoding, using the training samples $\mathcal{D}_{tr}$.}
In prediction, \rvc{they calculate} the center of the mapped vectors of few-shot images of each class, and \rvc{compare} a testing image to this center.
\rvc{Note learning the mapping of the class encoding impacts the learning of the mapping of the image features.}
Xiong et al. \cite{xiong2018one} \rvc{work} on one-shot KG completion with unseen relations. 
They \rvc{develop} a matching network to compare a testing entity pair with the one-shot entity pair of each unseen relation, where the features of an entity pair \rvc{(i.e., input encoding) are} learned by a neighbourhood encoder, and a matching score \rvc{is} predicted by an LSTM network.
%
Zhang et al. \cite{zhang2020fewa} \rvc{work on few-shot KG completion with unseen relations, with a similar idea as the above work \cite{xiong2018one}.
%
Zhao et al. \cite{zhao2020knowledge} jointly map the image features (input encoding) and the class encoding (which is based the fusion of KG embeddings and text embeddings)} into one common space by MLPs.
In prediction, a testing sample is compared with the few-shot samples of each unseen class via calculating the Consine similarity.
\rvc{As the method in \cite{li2019large}, the mapping learning of the class encoding impacts the mapping learning of the input encoding}.
Sui et al. \cite{sui2021knowledge} \rvc{propose} a KG-aware few-shot text classification method.
It \rvc{maps and compares a testing sample with the few-shot samples of each unseen class using a task-agnostic relation network and a task-relevant relation network armed with external knowledge from NELL}.
%
%
Zhang et al. \cite{zhang2021knowledge} \rvc{work} on few-shot relation extraction from text, utilizing concept-level \rvc{knowledge} from Wikidata \cite{vrandevcic2014wikidata} or UMLS \cite{mccray2003upper}.
They \rvc{match the mapped testing sample (i.e., an entity mention pair) to the mapped few-shot samples and to the mapped 
relation (class) encoding, and combine the two matching scores.
Note sample input mapping model is a network which considers the sentence features, the entity description features and the KG concept features, while the relation encoding is based on the relation representations extracted from the KG embeddings}.

\subsection{Data Augmentation Paradigm}

There have been some FSL studies that attempt to generate additional samples or sample features for the unseen classes by using KGs.
\rvc{We} divide these methods into two categories: \textit{Rule-based} and \textit{Generation-based}.
\rvc{Although rules (heuristics) can be directly applied to FSL by e.g., annotating labels to samples as in distant supervision, we only find one KG-aware FSL work of this category.}
Instead, we find some KG-aware FSL studies of generation-based, which usually utilize statistical generation models such as GANs \cite{goodfellow2014generative} and VAEs \cite{kingma2013auto}.
Next we will introduce the works of each category.

%
%
%

\rvg{\subsubsection{Rule-based}
\rvc{Tsai and Salakhutdinov \cite{tsai2017improving} use a simple heuristic rule for sample generation for one-shot image classification.
They take an attention mechanism over the class encodings, which are based on the embeddings of a KG extracted from WordNet, to select the most relevant unseen classes for a seen class, such that samples of the seen class are transformed into a set of quasi-samples of these unseen classes as additional training samples.}

\subsubsection{Generation-based}
Some works leverage GANs and VAEs to generate extra labeled data for the unseen classes conditioned on their auxiliary information, as the generation-based category in KG-aware ZSL.
For example, Wang et al. \cite{wang2019tackling} \rvc{work} on few-shot KG completion involving both unseen entities and unseen relations, and propose a triple generator with Conditional VAE \cite{sohn2015learning} to supplement the real triple set.
Following the basic idea of VAE, the encoder is implemented with a recognition network and a prior network to learn the variational posterior distribution $q_\theta(z|O)$ and the conditional prior distribution $p_\phi(z|o_r)$ by taking as input the embedded textual descriptions of the triples $O$ and that of the relations $o_r$, respectively.
Next, the decoder by a generative network is proceeded to reconstruct the triple embeddings ($g_h, g_r, g_t$) by sampling them from the latent semantics $z$ conditioned on $o_r$, i.e., $p_\varphi(g_h, g_r, g_t|z, o_r)$.
During testing, more triple embeddings of unseen entities and unseen relations can be generated conditioned on their latent semantics $z$.

Besides obtaining additional training data using generative models, there are also some other works that are motivated by the application of GANs in domain adaption. They attempt to generate features that are more transferable from the ``lots-of-samples'' domains to the different but related ``few-samples'' domains.
For example, Zhang et al. \cite{zhang2020relation} \rvc{work} out a general feature generation framework for addressing few-shot unseen relations in two tasks --- few-shot KG completion with unseen relations and few-shot relation extraction from text.
The framework is adversarially trained to generate the features that are invariant to the seen and unseen relations,} and transfer such features to unseen relations with weighted combination.
\rvg{The feature generation module is implemented by a CNN} to iteratively extract features from the entity pair (or from the text sentence for relation extraction) until the discriminator cannot distinguish features of the seen relations and the unseen relations.

\subsection{Propagation-based Paradigm}
\rvc{We find two KG-aware FSL studies that adopt the idea of model parameter propagation as in KG-aware ZSL but no studies adopting class belief propagation.}
This may be because the current methods usually focus on utilizing the few-shot samples.
\rvc{For few-shot KG completion tasks, graph propagation} is widely utilized for addressing unseen entities \rvc{or unseen relations} that have few-shot associated triples.
They often aggregate the embeddings of the neighbouring entities and relations, which are usually seen, to get the embedding of an unseen relation or entity.
\rvc{Figure \ref{fig:propagation} (b) shows this idea with an example.
The entity $e_0$, which is unseen without a trained embedding, is connected to seen entities $e_1$, $e_2$ and $e_3$ through some few-shot triples with relations of $r_1$, $r_2$ and $r_3$. With a propagation model, the embedding of $e_0$ can be predicted via the trained embeddings of $e_1$, $e_2$ and $e_3$.
We classify these methods into a new category named \textit{Embedding Propagation}.
We will next introduce the works of model parameter propagation and embedding propagation, mainly from two dimensions --- the propagation graph and the propagation model.}

\subsubsection{Model Parameter Propagation}
Peng et al. \cite{peng2019few} \rvc{work on WordNet augmented few-shot image classification.
They first do model parameter propagation as in KG-aware ZSL, which uses a GCN and a graph to predict classifier parameters of the unseen classes,
and then ensemble these predicted classifiers with the classifiers learned from the few-shot samples.}
Chen et al. \cite{chen2020knowledge} \rvc{use a graph whose nodes are seen and unseen classes, and whose edges are assigned by correlation weights between classes for few-shot image classification. They initialize the classifier parameters of each graph node (class), and then use a Gated Graph Neural Network (GGNN) to update the classifier parameters of each graph node with multiple iterations. The GGNN is trained with a cross-entropy loss on $\mathcal{D}_{tr}$ and $\mathcal{D}_{few}$ 
a regularisation term on the classifier parameters}.

\subsubsection{Embedding Propagation}
\rvc{The embedding propagation methods mainly aim at few-shot KG completion tasks which predict triples involving unseen entities or unseen relations, without re-training the embedding of the original KG. Thus the graph for propagation is the KG to complete itself.}
These unseen \gyx{entities (resp. relations)} are also named as \textit{out-of-KG \gyx{entities (resp. relations)}} in some papers since they are usually not observed in \rvc{the KG whose embeddings have been trained.}

As far as we know, Hamaguchi et al. \cite{hamaguchi2017knowledge} \rvc{propose} the earliest embedding propagation \rvc{method for KG completion with} unseen entities.
They \rvc{use} a GNN but \rvc{revise} its propagation mechanism for the KG, and \rvc{adopt} a translation-based objective function for scoring \rvc{a} triple and for a loss for training.
Wang et al. \cite{wang2019logic} \rvc{propose} a Logic Attention Network (LAN) to \rvc{get the embeddings of unseen entities from their neighbouring entities and relations.}
In LAN, logic rules are exploited to measure neighbouring relations’ usefulness, and neighbours connected by different relations have different weights \rvc{w.r.t.} an unseen entity.
Bhowmik and Melo \cite{bhowmik2020explainable} \rvc{use} a variant of Graph Transformer encoder to embed an unseen entity by aggregating its neighbours based on their relevance to a given relation. 
It predicts the object of a triple, and can explain the prediction by finding out paths from the subject to the object.
Ali et al. \cite{ali2021improving} \rvc{predict relations between seen entities and unseen entities, and between unseen entities. 
For predicting relations between unseen entities, they initialize the entity embeddings by the} entities' textual information using Sentence BERT, and then \rvc{propagate} to update the entities' embeddings by a graph encoder named StarE \cite{galkin2020message}. 

\rvc{Some simpler propagation models have also been explored for KG completion with unseen entities.
Besides GNN, Ali et al. \cite{ali2021improving} also explore a linear projection of entity features to relation features without considering the graph structure. This could also be classified as a mapping-based method by considering the entity features as input encoding and the relation features as class encoding}.
Dai et al. \cite{dai2020inductively} \rvc{use} two modules: an estimator which calculates a candidate set of embeddings for an unseen entity according to its all associated triples using the translation operation of TransE (or RotatE), and a reducer which calculates \rvc{the} unseen entity's embedding according to all its candidate embeddings.
%
Albooyeh et al. \cite{albooyeh2020out} \rvc{use} some simple aggregation operations such as averaging to get the embedding of an unseen entity from its neighbours.
\rvc{This method can be applied to any KG embedding models, but it requires that the original KG embedding training is adjusted for the aggregation operation.}
%

\rvc{All the above mentioned few-shot KG completion methods deal with unseen entities. We also find one embedding propagation method \cite{zhao2020attention} that can deal with both unseen entities and unseen relations.}
It mainly uses specific transition functions, aggregation functions and graph attention mechanisms to transform information from the associated triples to an unseen entity or relation, \rvc{where a triple is scored by a translation-based function and the model is trained with a margin loss. Note that this method does not deal with the situation with both unseen entities and unseen relations.}

\subsection{Class Feature Paradigm}
\rvc{The class feature paradigm of FSL is close to that of ZSL, as shown in  Figure \ref{fig:class_feature}.
As the mapping-based paradigm, many KG-aware ZSL methods of the class feature paradigm can be directly extended to support FSL by training $f'$ with both the training samples $\mathcal{D}_{tr}$ and the few-shot samples $\mathcal{D}_{few}$.
In this part, we focus on those class feature paradigm works that are originally proposed for FSL.
Some such works are found but not many since class feature fusion under the FSL setting does not significantly differ from class feature fusion under normal supervised learning settings.
We classify them into two categories: \textit{Text Feature Fusion} where the input encoding $g(x)$ and the class encoding $h(y)$ are both of text features, and \textit{Multi-modal Feature Fusion} where $g(x)$ and $h(y)$ are of different kinds of features.
Next we will introduce the works of each category from the dimensions of input encoding, class encoding and the fusion model.
}

\subsubsection{Text Feature Fusion}
\rvc{We find one KG-aware FSL work that belongs to text feature fusion.
Banerjee et al. \cite{banerjee2020self} propose a KG-aware method for few-shot question answering.
The input (the text context and the question) and the class (answer) are fed into a transformer-based model as $f'$.
This model is learned by span masked language modeling from KG triples, each of which simulate a combination of context, question and answer.
}
Note that the method can also support ZSL as introduced in Section \ref{sec:zsl_class_feature}.

\subsubsection{Multi-modal Feature Fusion}
\rvc{Fusing input encoding and class encoding of different kinds is harder and often requires a more complicated fusion model}.
We find two KG-aware FSL studies of this category. 
Zhang et al. \cite{zhang2019long} \rvc{investigate} text relation extraction for long-tailed relations which have few-shot samples (sentences).
The proposed method uses a GCN to learn the embedding of each relation \rvc{as class encoding, over a KG \gyx{derived from Freebase} where  the hierarchical relationship between relations are modeled, and then feeds the class encoding and the sentence features (input encoding) into an attention-based model}.
Yang et al. \cite{yang2021empirical} \rvc{investigate} few-shot visual question answering. 
A baseline \rvc{that they adopt, named KRISP \cite{marino2021krisp}, uses KGs such as ConceptNet for augmentation and is applied to this task.
In KRISP, the features of the image and the question are first fused by a Transformer-based model for input encoding, and then the input encoding is further fused with the class encoding (features of knowledge retrieved from the KG) to predict the answer.}

\subsection{Optimization-based Paradigm}
\rvc{To optimize the training with the few-shot samples $\mathcal{D}_{few}$,
some meta-learning algorithms have \rvc{been} applied for} fast adaption and for avoiding over-fitting by obtaining better initial parameter settings, more optimized searching steps or more suitable optimizers.
\rvg{We regard these FSL methods as \textit{Optimization-based Paradigm} and present their general workflow in Figure \ref{fig:optimization}.}
\rvg{To mimic the learning with the few-shot samples that are assumed to be available in testing, they all adopt an episode-based training strategy and generate a set of learning tasks.
Each task consists of a support set and a query set to simulate the few-shot sample set $\mathcal{D}_{few}$ and the testing samples $\mathcal{D}_{te}$, respectively, and aims at learning a Meta Learner parameterized by $\theta$, which is expected to be able to learn efficiently from only a small number of samples. 
After multiple training iterations on these training tasks, the method can obtain an optimal Meta Learner.
More formally, in a training task $t$, we have the support set $\mathcal{D}_t^{support}$ and the query set $\mathcal{D}_t^{query}$, the Meta Learner $F_\theta$ can either automatically produce a model $f$ from $\mathcal{D}_t^{support}$ to predict the samples in $\mathcal{D}_t^{query}$, or learn optimized gradients $\nabla_f$ from $\mathcal{D}_t^{support}$ to better update the model parameters on $\mathcal{D}_t^{query}$. 
The optimization at each step $t$ is generated by computing the loss on the query set $\mathcal{D}_t^{query}$.
When at test time, a set of tasks disjoint with the training tasks is designed for evaluation, where the support set is from $\mathcal{D}_{few}$, and the query set is $\mathcal{D}_{te}$ we need to predict.
}

\rvc{We collect} quite a few KG-aware FSL methods of this paradigm: some of them are for KG completion tasks with unseen entities or relations that are associated with a small number of triples \cite{chen2019meta,wang2019meta,baek2020learning,zhang2020fewa,zhang2020fewb, lv2019adapting}, while the others are for KG augmented few-shot image classification and few-shot text relation extraction \cite{qu2020few,zhang2019tgg,zhang2021knowledge}.
We find some studies develop new meta learning algorithms or revise the existing ones w.r.t. the KG, while some other studies just apply meta learning independently without specifically considering the KG context.
We thus classify \rvc{these FSL methods} into \textit{KG-specific Optimization} and \textit{KG-agnostic Optimization}.

\rvc{
\begin{figure}
\centering
\includegraphics[width=0.5\textwidth]{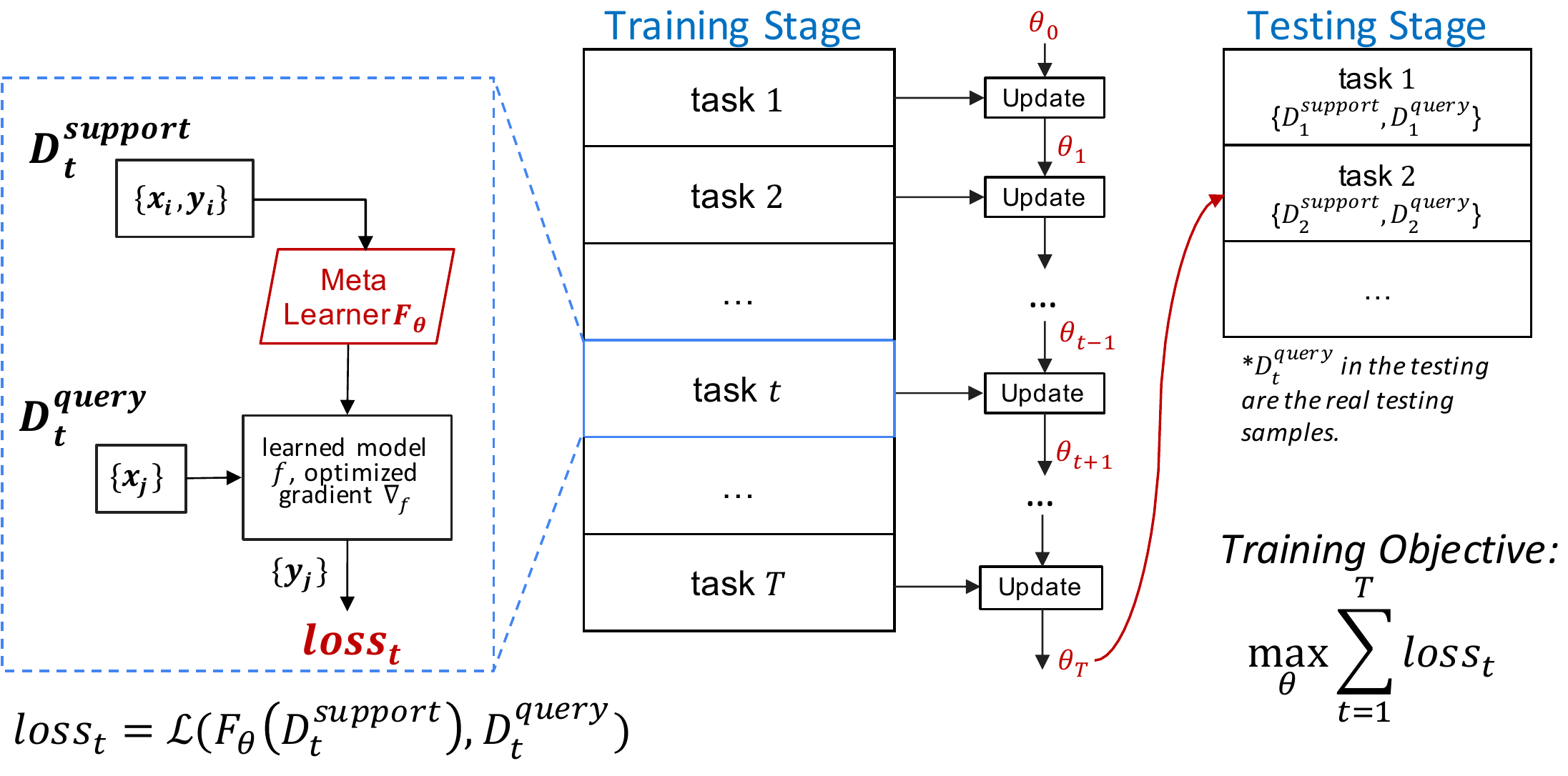}
\caption{\rvg{Overview of the optimization-based paradigm of FSL}. \label{fig:optimization}}
\end{figure}
}

\subsubsection{KG-specific Optimization}
%
\rvg{Chen et al. \cite{chen2019meta} propose a new method named MetaR to predict triples involving unseen relations which have only a small number of associated triples.
A learning task is defined for a specific relation, and each sample is its one associated entity pair and the number of support samples is set to be no more than $5$.
Firstly, a relation-meta learner is designed to extract higher-order relation representations from the embeddings of the support entity pair as relation meta, and then the gradient meta, which will guide how the relation meta should be efficiently updated, is generated by feeding the relation meta and each entity pair into an embedding learner to compute the triple score.
Finally, the updated relation meta is transferred to the triples in the query set to compute their scores via the same embedding learner.
The loss of query set is used to update the whole model so as to quickly learn better relation meta for testing when only a small number of support samples are given.}

\rvc{Meanwhile the typical meta learning algorithm Model-Agnostic Meta-Learning (MAML) which is to learn a good parameter initialization for a new meta-learning task \cite{finn2017model} is often adopted and augmented with KG.}
Wang et al. \cite{wang2019meta} \rvc{work} on a few-shot KG reasoning task which is to predict the tail entity given a head entity and an unseen relation and infer paths from the head entity to the tail entity.
They \rvc{augment} MAML with additional task (relation) specific information encoded by a neighbour encoder based on embedding concatenation and linear transformation operations, and a path encoder based on LSTM.
Baek et al. \cite{baek2020learning} \rvc{work} on a realistic few-shot KG completion task, where relations between seen entities and unseen entities, and between unseen entities are both predicted using GNNs.
\rvg{They propose a Graph Extrapolation Network for quickly learning the embeddings of unseen entities with only a few associated triples, where a set of tasks are formulated with simulated unseen entities so as to generalize to the real unseen entities raised at test time.}

\subsubsection{KG-agnostic Optimization}
\rvc{In some other optimization-based FSL studies, the application of meta learning algorithms is independent of the KG.
Note that some of these studies are still reviewed as they aim at KG related prediction tasks.}
Lv et al. \cite{lv2019adapting} \rvc{work on the same task as \cite{wang2019meta}, i.e., few-show KG completion} with unseen relations.
They \rvc{adopt} reinforcement learning to search \rvc{tail entities and paths which could infer these tail entities}, and directly \rvc{apply} MAML with one relation modeled as one task.
Zhang et al. \cite{zhang2020fewb} \rvc{propose another method for few-shot KG completion} with unseen relations, where MAML is directly applied for well initializing an on-policy reinforcement learning model for fast adaption.
Qu et al. \cite{qu2020few} \rvc{work} on few-shot relation extraction by modeling the posterior distribution of prototype vectors for different relations.
To this end, they first \rvc{initialize} the relation prototype vectors by a BERT model over the samples (i.e., sentences) and a GNN over a global relation graph extracted from different ways, and then effectively learn their posterior distribution by a Bayesian meta-learning method which is related to MAML but can handle the uncertainty of the prototype vectors.

It is worth mentioning that meta \rvc{learning-based optimization can simply act as a complement for model training} in methods of other paradigms.
Zhang et al. \cite{zhang2020fewa} \rvc{predict} KG triples with unseen relations.
Their few-shot relational learning method FSRL, \rvc{which is of the mapping-based paradigm as it compares a testing entity pair with few-shot samples of each unseen relation after mapping, uses MAML for fast adaption in training the mapping models}.
Zhang et al. \cite{zhang2019tgg} \rvc{attempt} to address both zero-shot and few-shot image classification, with an approach named Transfer Graph Generation (TGG) which has a graph generation module for generating instance-level graph, and a propagation module for utilizing this graph for prediction. 
They \rvc{train} the whole model with an episodic training strategy of meta learning.
Zhang et al. \cite{zhang2021knowledge} \rvc{use} a joint mapping method to predict relations for entity mentions in a sentence.
In this method, a knowledge-enhanced prototypical network and a relation meta learning model, which implement the matching between instances and the matching between instance and relation meta, respectively, are trained with gradient meta.

\subsection{Transfer-based Paradigm}\label{sec:tbp}
Some KG-aware FSL methods directly apply models that are built \rvc{from samples} of seen classes ($\mathcal{D}_{tr}$) to predicting \rvc{samples} of unseen classes ($\mathcal{D}_{te}$) with the help of the few-shot samples ($\mathcal{D}_{few}$).
These methods are \rvc{regarded as} the transfer-based paradigm. 
It is worth noting that \rvc{some methods of the other paradigms such as the model parameter propagation methods} also have an idea of implicitly transferring data or parameters from seen classes to unseen classes.
\rvc{The difference is that methods of the transfer-based paradigm  directly apply the whole prediction models learned from $\mathcal{D}_{tr}$ to $\mathcal{D}_{te}$.}
\rvc{They are often for few-show KG completion, where one KG is given for model learning, while another KG composed of triples of unseen entities is for inference (prediction), as shown in Figure \ref{fig:transfer}}.
For convenience, we name the first KG as the seen KG and the second KG as the unseen KG.
Such a task in common in real-world: \rvc{given an existing KG (seen), the unseen KG could be either an emerging sub-KG that is to be added, or a KG of another domain with shared relations but different entities.
We} further classify these FSL methods into two categories: \textit{Neural Network Transfer} and \textit{Rule Transfer}.
\rvc{We will next introduce the works of each category mainly from the dimension of the specific model to transfer.}

\rvc{
\begin{figure}
\centering
\includegraphics[width=0.48\textwidth]{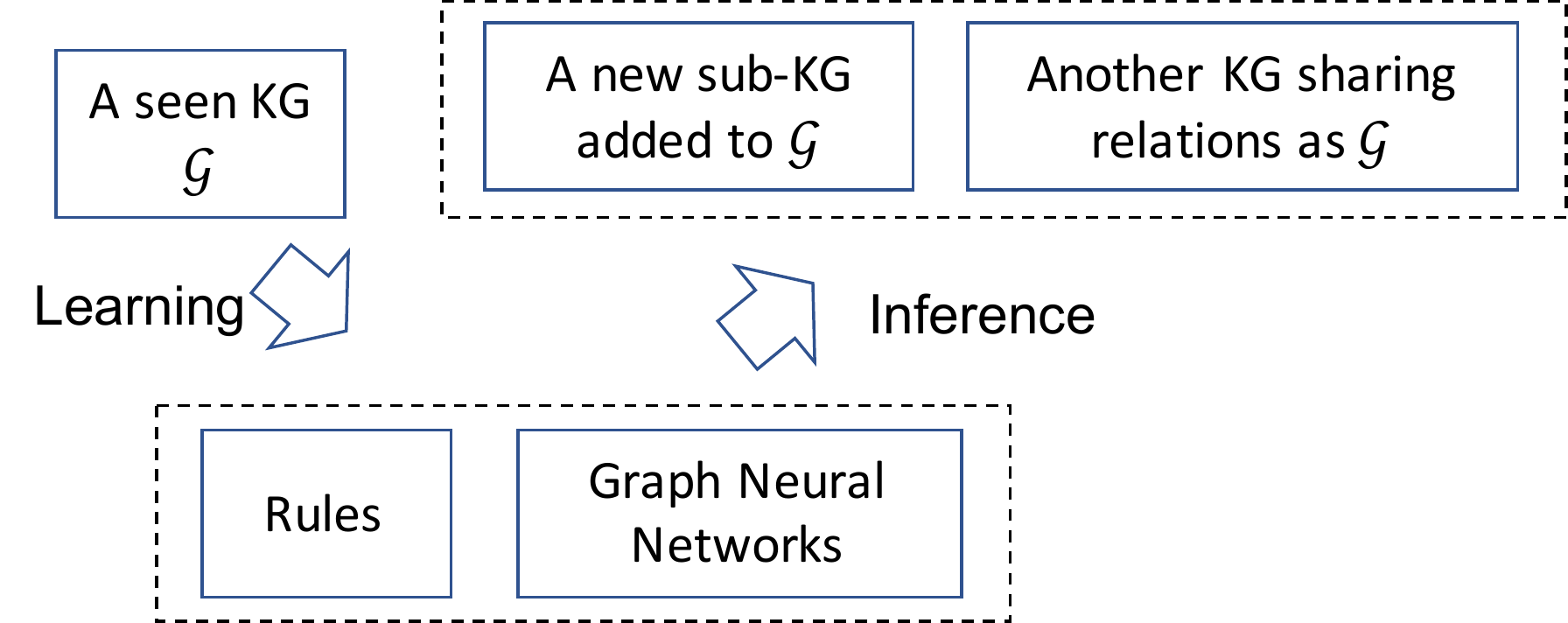}
\caption{\rvc{Overview of the transfer-based paradigm of FSL when applied to KG completion}. \label{fig:transfer}}
\end{figure}
}

\subsubsection{Neural Network Transfer}
\rvc{As GNNs can represent statistical regularities and  structure  patterns in a graph,
some methods transfer a GNN learned from the seen KG to the unseen KG for inference.}
Teru et al. \cite{teru2020inductive} \rvc{propose} a method named GraIL. It learns a GNN by extracting subgraphs from the seen KG and labeling their entities with their structural roles (e.g., the shortest distance between two entities), and apply this GNN to predict the relation between two unseen entities in the unseen KG.
Chen et al. \cite{chen2021topology} \rvc{extend} GraIL by using R-GCN \cite{schlichtkrull2018modeling} for supporting multiple relations in the KG. 
\rvc{Besides, they propose to transfer another model named Relational Correlation Network learned from the seen KG to the unseen KG, and combine its triple score with that by the extended GraIL.
Note that the relational correlation network is learned from a relation correlation graph}
whose nodes represent the relations and whose edges indicate the topological correlation patterns between any two relations in the original KG.
Liu et al. \cite{liu2021indigo} \rvc{propose to reformulate the original KG as a graph as follows:} 
two connected KG entities or an entity and its own, are represented as one graph node, and each node is initialized with features indicating the triples in which the two entities are involved.
\rvc{They learn} a GCN from the graph of the seen KG, 
which is shown to be able to capture \rvc{graph patterns represented in Datalog rules, and apply this GCN  to predict graph node features (i.e., triples) of the unseen KG.}

\subsubsection{Rule Transfer}
\rvc{Different rules such as Horn rules, first-order rules and} their weighted versions can be learned from a KG for represent graph patterns and regularities \cite{galarraga2013amie,kimmig2012short}.
They may not be as \rvc{expressive as GNNs} for representing very complicated statistical regularities, but are more interpretable.
Sadeghian et al. \cite{sadeghian2019drum} \rvc{propose a method named DRUM for few-shot KG completion}, where first-order logical rules (such as $brother(X,Z) \land fatherOf(Z,Y) \rightarrow uncleOf(X,Y)$) associated with weights are learned from the seen KG by a differentiable way using the rule mining method named Neural LP, and these rules are applied in the unseen KG for deductive reasoning for new triples.
This method uses the KG relations as the rule predicates and assumes that the relations of the seen and the unseen KGs are the same, such that the rules can be directly transferred.
\rvc{For the situation where predicates (relations) of the unseen KG are different from those of the seen KG, we find the following two solutions for rule transfer}: \textit{(i)} matching predicates between rules, proposed by Mihalkova et al. \cite{mihalkova2007mapping,mihalkova2009transfer} \rvc{for transferring} rules mined by  Markov Logic Networks (MLNs), and \textit{(ii)} extracting and transferring \rvc{more general higher order rules that are summarized from the original rules \cite{davis2009deep,van2015todtler}.}

\rvc{
\subsection{ZSL and FSL Comparison}

Some KG-aware FSL methods are specifically developed to utilize the few-shot samples. 
Typical kinds of such methods are the optimization-based paradigm, the transfer-based paradigm, and the embedding propagation category of the propagation-based paradigm.
The meta learning algorithms used in the optimization-based paradigm and the models directly transferred both rely on the few-shot samples, and thus cannot be applied to ZSL.
For example, when a set of rules or a GNN are transferred to predict triples involving unseen entities, these unseen entities must be associated with some triples for evidences (graph patterns) for inference.
Many of these FSL methods aim at KG completion tasks. They often ignore or do not well utilize the auxiliary information.

Meanwhile, some other KG-aware FSL methods are simple extensions of corresponding KG-aware ZSL methods. They train the original ZSL models with the additional few-shot samples, or ensemble the ZSL models with the models trained from the few-shot samples.
This is common in methods of the mapping-based paradigm and the class feature paradigm.
The mapping models and the fusion models can be trained with both the training samples and the few-shot samples (e.g., \cite{ma2016label,akata2015label,rios2018few}).
Actually, the majority of the ZSL methods, which usually well utilize the auxiliary KG, can be extended to support FSL with the above extension ideas, although for the class belief propagation category, there are currently only ZSL works but no FSL works. However, well combing the KG (or some other auxiliary information) with the few-shot samples is still an open problem.

}

\section{Applications and Resources}\label{sec:app}
In this section, we first very briefly revisit \rvc{KG-aware ZSL and FSL in different domains and tasks, and then introduce some public resources. See Table \ref{table:taskworks} for a summary of methods of each task, and see Figure \ref{fig:resource} for an overview of benchmarks of each task}.

\rvc{
\begin{figure*}
\centering
\includegraphics[width=0.9\textwidth]{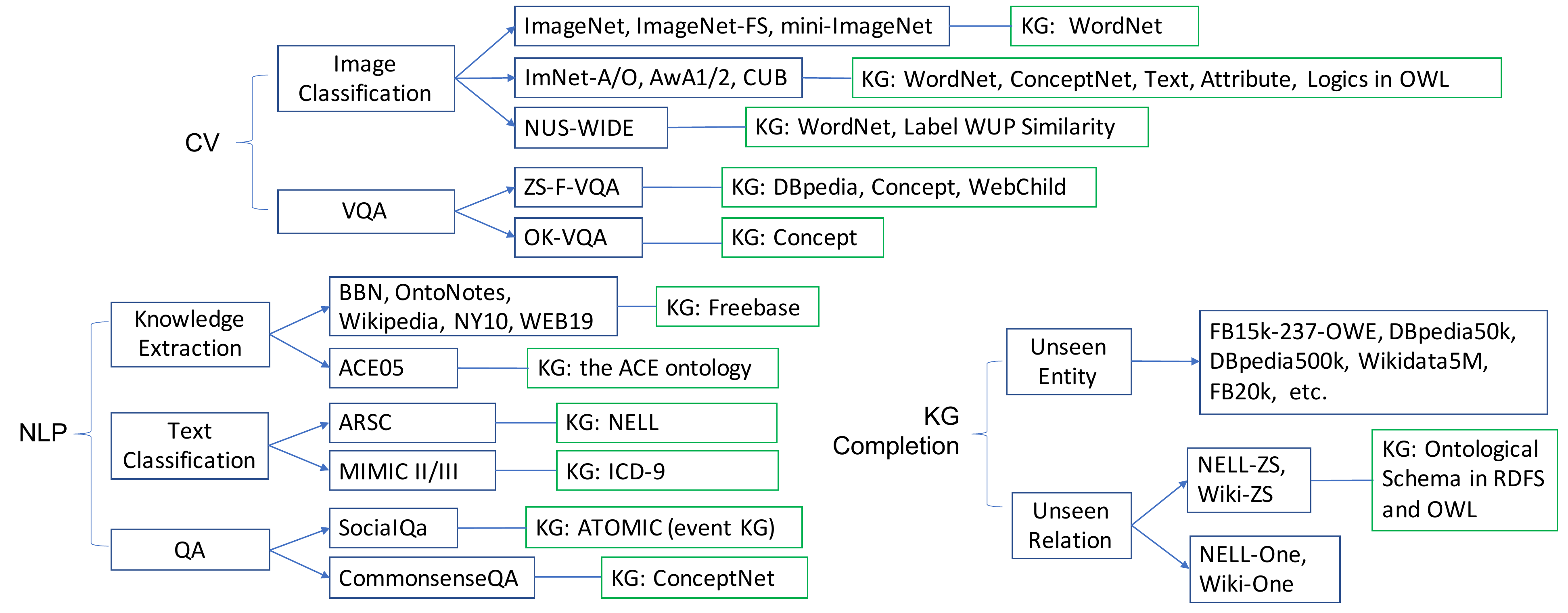}
\caption{\rvc{Overview of the benchmarks of different tasks and their KGs. The green box denotes the KG auxiliary information. Some KG completion benchmark without additional KG (ontology) auxiliary information are also collected, as the zero-shot or few-shot task itself is within a KG context.} \label{fig:resource}}
\end{figure*}
}

\begin{table*}[t]
\footnotesize{
\centering
\renewcommand{\arraystretch}{1.3}
\begin{tabular}[t]{m{2.5cm}<{\centering}|m{2.8cm}<{\centering}|m{5cm}<{\centering}<{\centering}|m{5cm}<{\centering}}\hline
 \textbf{Tasks} &\textbf{Paradigms} &\textbf{ZSL Works} & \textbf{FSL Works}   \\ \hline
\multirow{4}{*}{Image Classification}   & Mapping-based & \cite{palatucci2009zero,akata2013label,akata2015label,li2015zero,changpinyo2016synthesized,liu2018combining,nayak2020zero,li2020transferrable,roy2020improving,chen2020ontology} & \cite{akata2013label,akata2015label,li2019large,zhao2020knowledge,li2020transferrable,jayathilaka2021ontology,monka2021learning}  \\ \cline{2-4}
   & Data Augmentation & \cite{zhang2019tgg,geng2021ontozsl} & \cite{tsai2017improving} \\ \cline{2-4}
   & Propagation-based & \cite{lee2018multi,wang2018zero,kampffmeyer2019rethinking,wei2019residual,chen2020zero,luo2020context,geng2020explainable,wang2021zero} & \cite{peng2019few,chen2020knowledge} \\ \cline{2-4}
   & Optimization-based & --- & \cite{zhang2019tgg} \\ \hline
\multirow{2}{*}{VQA} & Mapping-based & \cite{chen2020ontology,chen2021zero} &--- \\ \cline{2-4}
 & Class Feature & --- &\cite{marino2021krisp} \\ \hline
\multirow{4}{*}{Knowledge Extraction} & Mapping-based & \cite{ma2016label,huang2018zero,imrattanatrai2019identifying,li2020logic} & \cite{ma2016label} \\ \cline{2-4}
&Data Augmentation & --- & \cite{zhang2020relation} \\ \cline{2-4}
 & Class Feature & \cite{nguyen2021dozen,ristoski2021kg,gong2021prompt} &\cite{zhang2019long} \\ \cline{2-4}
 & Optimization-based & --- & \cite{qu2020few,zhang2021knowledge} \\ \hline
 \multirow{2}{*}{Text Classification} & Mapping-based & \cite{rios2018few,chen2021zero} & \cite{rios2018few,sui2021knowledge} \\ \cline{2-4}
 & Class Feature & \cite{zhang2019integrating} & --- \\ \hline
\multirow{2}{*}{QA} & Propagation-based & \cite{bosselut2021dynamic} & --- \\ \cline{2-4}
 & Class Feature & \cite{banerjee2020self,zhou2021encoding} &\cite{banerjee2020self} \\ \hline
\multirow{6}{*}{\thead{KG Completion \\ (Unseen Entity)}} & Mapping-based & \cite{shah2019open,hao2020inductive} & --- \\ \cline{2-4}
& Data Augmentation & --- & \cite{wang2019tackling} \\ \cline{2-4}
& Propagation-based & --- & \cite{hamaguchi2017knowledge,wang2019logic,bhowmik2020explainable,albooyeh2020out,zhao2020attention,dai2020inductively} \\ \cline{2-4}
 & Class Feature & \cite{zhao2017zero,shi2018open,yao2019kg,wang2021inductive,zha2021inductive,wang2021structure,niu2021open,wang2021kepler,amador2021ontology,ali2021improving} & --- \\ \cline{2-4}
 &Optimization-based & --- & \cite{baek2020learning} \\ \cline{2-4}
 &Transfer-based & --- & \cite{sadeghian2019drum,teru2020inductive,liu2021indigo,chen2021topology} \\ \hline
\multirow{5}{*}{\thead{KG Completion \\ (Unseen Relation)}} & Mapping-based & --- & \cite{xiong2018one,zhang2020fewa} \\ \cline{2-4}
& Data Augmentation & \cite{rocktaschel2015injecting,qin2020generative,geng2021ontozsl} & \cite{wang2019tackling,zhang2020relation} \\ \cline{2-4}
& Propagation-based & --- & \cite{zhao2020attention} \\ \cline{2-4}
 & Class Feature & \cite{yao2019kg,zha2021inductive,wang2021structure} & --- \\ \cline{2-4}
 &Optimization-based & --- & \cite{wang2019meta,chen2019meta,lv2019adapting,zhang2020fewb} \\ \hline
\end{tabular}
\vspace{0.1cm}
\caption{\rvc{A summary of KG-aware ZSL and FSL works of different tasks}.}\label{table:taskworks}
}
\end{table*}

\subsection{Computer Vision}

\subsubsection{Image Classification} 

Regarding zero-shot image classification\footnote{\rvc{Note that object recognition is often transformed into two steps: discovering object bounding boxes and classifying these bounding boxes. Zero-shot object recognition in many works (e.g., \cite{li2019zero}) usually aim at the second step and are equivalent to zero-shot image classification. There are also some works using KGs for supporting object recognition \cite{fang2017object,lang2021contrastive}.}}, the early \rvc{works mainly utilize class attributes \cite{lampert2009learning,lampert2013attribute,farhadi2009describing, parikh2011relative} and class text information \cite{norouzi2014zero, socher2013zero,frome2013devise,elhoseiny2013write,qiao2016less,reed2016learning}, with the mapping-based paradigm and the data augmentation paradigm often adapted.}
However, the state-of-the-art performance on many \rvc{benchmarks} now are achieved by those methods utilizing KGs constructed by various sources including existing KGs, \rvc{task-specific data and domain knowledge \cite{zhang2019tgg,roy2020improving,nayak2020zero,wang2018zero,kampffmeyer2019rethinking,geng2021ontozsl}}.
To utilize the KGs, the propagation-based paradigm starts to be widely adopted in some recent studies such as \cite{wang2018zero,kampffmeyer2019rethinking,geng2020explainable}.

To support method development and evaluation, some open benchmarks on KG-aware zero-shot image classification have been proposed:
\begin{itemize}
\item \textbf{ImageNet} \rvc{is a large-scale image database containing a total of 14 million images from 21K classes \cite{deng2009imagenet}.}
Each image is labeled with one class, each class is matched to a WordNet \cite{miller1995wordnet} entity, and the class hierarchies from WordNet can be used as the auxiliary information.
\rvc{In the works \cite{wang2018zero} and \cite{kampffmeyer2019rethinking}}, 1K classes with balanced images are used as seen classes for training, while classes that are 2-hops or 3-hops away, or all the other classes are used as unseen classes for testing.
%
The weakness of ImageNet mainly lies in that the \rvc{KG has only class hierarchies and class name)} without any other knowledge such as class attributes and commonsense knowledge.

\item \textbf{ImNet-A} and \textbf{ImNet-O} are extracted from ImageNet by Geng et al. \cite{geng2021ontozsl,geng2023benchmarking}. ImNet-A includes $80$ classes from $11$ animal species, while ImNet-O including $35$ classes of general objects. In the experiment in \cite{geng2021ontozsl}, ImNet-A is partitioned into 28 seen classes (37,800 images) and 52 unseen classes (39,523 images), while ImNet-O is partitioned into 10 seen classes (13,407 images) and 25 unseen classes (25,954 images).
In their latest version released in \cite{geng2023benchmarking}, each benchmark is equipped with a KG which is semi-automatically constructed with several kinds of auxiliary knowledge, including class attribute, class textual information, commonsense knowledge from ConceptNet, class hierarchy (taxonomy) from WordNet and logical relationships such as disjointness.

\item \textbf{AwA2}, originally proposed in \cite{xian2018zero}, \rvc{has $50$ animal classes, $37, 322$ images collected from public Web sources such as Flickr and Wikipedia, and $85$ real-valued attributes annotated by experts} for describing animal visual characteristics.
It can also be used to evaluate KG-aware ZSL methods, since the classes are aligned with WordNet entities and the animal taxonomy from WordNet can be used as a simple KG.
In the extended version by Geng et al. \cite{geng2023benchmarking}, a KG is constructed for AwA2 with the same types of knowledge as ImNet-A and ImNet-O.
Note AwA in \cite{geng2023benchmarking} actually refers to AwA2, while the original AwA1 released in \cite{lampert2013attribute} does not have public copyright license for \rvc{all of} its images.

\item \textbf{NUS-WIDE} \cite{chua2009nus} \rvc{is} a multi-label image classification dataset including nearly \rvc{$270$K} images.
\rvc{Each image contains multiple objects, and thus NUS-WIDE is widely used for evaluating multi-label zero-shot image classification} \cite{lee2018multi,huang2020multi,narayan2021discriminative}.
To be more specific, the images have two versions of label sets: \rvc{NUS-1000 and NUS-81. The former comprises $1000$ noisy labels collected from Flickr user tags and the latter is a dedicated one with $81$ human-annotated labels.}
To perform multi-label ZSL, the labels in NUS-81 is taken as the unseen label set, while the seen label set is derived from NUS-1000 with $75$ duplicated ones removed and thus results in $925$ seen label classes.
In \rvc{works on KG-aware ZSL} such as \cite{lee2018multi}, NUS-WIDE is accompanied by a KG with $3$ types of label relations, including a super-subordinate correlation from WordNet, positive and negative correlations computed by label similarities such as WUP similarity \cite{wu1994verb}.

\end{itemize}

For few-shot image classification, the majority of the existing methods aim at utilizing the few-shot samples by e.g., meta learning, while the KG-aware studies often try to combine benefits from the KG external knowledge and the few-shot samples.
Some of them simply extend their mapping-based models which are originally developed for zero-shot image classification by training with additional samples of the unseen classes (e.g., \cite{jayathilaka2021ontology,akata2015label,li2019large}), while some others further generate more data for unseen classes conditioned on KGs (e.g., \cite{tsai2017improving}) or utilize KGs to transfer images features from seen classes to unseen classes (e.g., \cite{chen2020knowledge,peng2019few}).

There are also some open benchmarks that can be used for KG-aware few-shot image classification. The following are some widely used \rvc{ones}:
\begin{itemize}
    \item \textbf{ImageNet-FS} \cite{hariharan2017low} and \textbf{mini-ImageNet} \cite{vinyals2016matching} are two derivatives of ImageNet.
    ImageNet-FS covers $1,000$ ImageNet classes with balanced images and these classes are divided into $389$ seen classes and $611$ unseen classes. During evaluation, images of $193$ seen classes and $300$ unseen classes are used for cross validation, while images of the remaining $196$ seen classes and $311$ unseen classes are used for testing.
    in contrast, mini-ImageNet is relatively small. It has $100$ classes, each of which has $600$ images. These classes are partitioned into $80$ seen classes and $20$ unseen classes.
    \rvc{Since all the classes are aligned with WordNet entities, WordNet 
    can be used as the external knowledge.}
    
    \item \textbf{AwA1} \cite{lampert2013attribute}, \textbf{AwA2} \cite{xian2018zero} and \textbf{CUB} \cite{wah2011caltech} are three typical zero-shot image classification benchmarks that can be easily extended for a few-shot setting. AwA1 and AwA2 both have $50$ coarse-grained animal classes, with $40$ of them being seen classes and the remaining being unseen classes.
    CUB has $200$ fine-grained bird classes, with $150$ of them being seen classes and the remaining be unseen classes.
    A small number of labeled images (usually $10$) are added for each unseen class so as to support a few-shot setting.
    Meanwhile, several KGs have been added to these benchmarks for evaluating KG-aware methods: Tsai and Salakhutdinov \cite{tsai2017improving} and Akata et al. \cite{akata2013label,akata2015label} \rvc{add} WordNet classes hierarchies to AwA1 and CUB; Zhao et al. \cite{zhao2020knowledge} \rvc{construct} a domain-specific KG for CUB based on the attribute annotations of samples;
    Zhang et al. \cite{zhang2019tgg} \rvc{exploit} ConceptNet to construct a KG for AwA2, and \rvc{utilize} part-level attributes to construct a KG for CUB.
\end{itemize}

\subsubsection{Visual Question Answering (VQA)}
\rvc{VQA is to answer a natural language question according to a given image}.
Teney et al. \cite{teney2016zero} ﬁrst \rvc{propose zero-shot VQA as the setting where there are unseen concepts in the text}. Namely, a testing sample \rvc{is} regarded as unseen if there \rvc{is} at least one novel word in its question or answer.
Ramakrishnan et al. \cite{ramakrishnan2017empirical} \rvc{consider} novel objects in the image. Namely, an image object that \rvc{has} never appeared in the training images \rvc{is} regarded as unseen.
%
KGs have been exploited for addressing zero-shot VQA, but not widely.
\rvc{The work \cite{chen2021zero} proposes a mapping-based method, where answers that have never appeared in training are predicted via comparing the KG-based embeddings of the question and answer embeddings. 
%
The work \cite{chen2020ontology} also adopts the mapping-based paradigm, but builds and embeds} an OWL ontology for establishing connections between seen answers and unseen answers.

A few VQA datasets have been published, but only a small number of them have been used \rvc{for KG-aware zero-shot VQA}:
\begin{itemize}
\item \textbf{ZS-F-VQA} \cite{chen2021zero}, constructed by re-splitting a fact-based VQA benchmark named F-VQA \cite{wang2017fvqa}, has no overlap between answers of the training samples and answers of the testing samples. 
In average, the training set has $2,384$ questions, $1,297$ images, and $250$ answers, while the testing set has $2,380$ questions, $1,312$ images, and another $250$ answers.
Chen et al. \cite{chen2021zero} \rvc{extract} facts from three public KGs (DBpedia \cite{auer2007dbpedia}, ConceptNet \cite{speer2017conceptnet} and WebChild \cite{tandon2014webchild}), and \rvc{construct a auxiliary KG} for evaluating KG-aware methods.

\item \textbf{OK-VQA} \cite{marino2019ok} is a recent benchmark where the visual content of an image is not sufficient to answer the question.
It has $14,031$ images and $14,055$ questions, and the correct answers are annotated by volunteers.
Chen et al. \cite{chen2020ontology} \rvc{use} it for evaluating KG-aware zero-shot VQA, by extracting $768$ seen answers and $339$ unseen answers, using \rvc{auxiliary information} from ConceptNet.
\end{itemize}

Regarding few-shot VQA, the existing methods \rvc{(e.g., \cite{yang2021empirical})} often rely on pre-trained language models such as GPT-3 which have already learned a large quantity of knowledge from text corpora.
To incorporate images, visual language models can be pre-trained with images and text, or images can also be transformed into text by e.g., image captions so as to be utilized in language models \cite{banerjee2021weaqa}.
\rvc{Meta learning is also applied for fully utilizing the few-shot samples and fast model training} \cite{teney2018visual}.

KGs \rvc{have} complementary knowledge besides the pre-trained (visual) language models and the few-shot samples. \rvc{We find some KG-aware few-shot VQA studies but no open benchmarks.}
Yang et al. \cite{yang2021empirical} \rvc{propose} a supervised learning method which \rvc{use} knowledge retrieved from KGs for augmenting the question-answer samples, and this method \rvc{is} used as a baseline in comparison with the GTP-3-based method.
Marino et al. \cite{marino2021krisp} first \rvc{fuse} features of the question and the image by a Transformer-based model, and then \rvc{fuse} these features with knowledge from ConceptNet.
\rvc{On the other hand}, the aforementioned mentioned zero-shot VQA benchmarks ZS-F-VQA and OK-VQA can be easily adjusted by adding few-shot samples for \rvc{supporting} the few-shot VQA setting.

\subsection{Natural Language Processing}
\subsubsection{Knowledge Extraction}
By knowledge extraction, we refer to those NLP tasks that are to extract structured \rvc{knowledge} including entities, relations, events and so on from natural language text. Relational facts, which are sometimes simply called triples in this domain, can also be extracted, after entities and relations are recognized.
Since the entities, relations \rvc{and} events can often be aligned with elements in a KG (such as a general purpose KG and an event ontology), their relationships represented in the KG can be exploited to address both zero-shot and few-shot \rvc{knowledge extraction.
For these tasks, most KG-aware zero-shot methods follow} the mapping-based paradigm utilizing the entities', relations' or events' embeddings in the KG \cite{huang2018zero,ma2016label,imrattanatrai2019identifying,li2020logic}, while KG-aware few-shot methods often follow the optimization-based paradigm \rvc{which utilizes} meta learning algorithms for fast training \cite{qu2020few,zhang2021knowledge}.
\rvc{Some methods also consider the class feature paradigm by fusing features from a KG with the input features for both ZSL and FSL \cite{nguyen2021dozen,zhang2019long}}. 

\rvc{There are quite a few knowledge extraction benchmarks that can be used for evaluating both KG-aware ZSL and KG-aware FSL.
Here are several representative ones:}

\begin{itemize}

    \item \textbf{BBN}, \textbf{OntoNotes} and \textbf{Wikipedia} are three benchmarks for fine-grained named entity typing, where the entity types are (partially) matched with types in Freebase.
    They are all adopted by Ma et al. \cite{ma2016label} for evaluating zero-shot entity typing, where the training set has only coarse-grained types, while the testing set has the second-level (fine-grained) types.
    They \rvc{use} a set of manually annotated documents (sentences) for validation and testing with a partitioning ratio of 1:9.
    Specifically, BBN has $2,311$ manually annotated Wall Street Journal articles with around $48$K sentences and $93$ two-level hierarchical types \cite{weischedel2005bbn}. 
    $47$ out of $93$ types are mapped to Freebase types.
    $459$ documents ($6.4$K sentences) are used for validation and testing.
    OntoNotes is an incrementally updated corpus that covers three languages (English, Chinese, and Arabic) and four genres (NewsWire, Broadcast News, Broadcast Conversation, and Web text) \cite{Weischedel2017OntoNotesA}. 
    It has $13,109$ news documents that are manually annotated using $89$ three-level hierarchical types. 
    $76$ manually annotated documents ($1,300$ sentences) are used for validation and testing.
    Wikipedia has around $780.5$K Wikipedia articles ($1.15$M sentences), $112$ fine-grained Freebase type annotations, \rvc{and $434$ validation and testing sentences}.

    \item \textbf{NYT10} and \textbf{WEB19} are two benchmarks used in \cite{imrattanatrai2019identifying} for zero-shot relation (property) extraction.
    NYT10 is constructed by Freebase triples and New York Times (NYT) corpus \cite{riedel2010modeling}. 
    WEB19 is formed by first selecting predicate paths in the FB15k benchmark \cite{bordes2013translating} as properties, then generating samples (a text corpus) associated with these properties using Microsoft Bing search engine API with the aid of human evaluation \cite{imrattanatrai2019identifying}. Under the ZSL setting in \cite{imrattanatrai2019identifying}, $217$ and $54$ properties of WEB19 are set to seen (for training) and unseen (for validation and testing), respectively, while a ll of the $54$ properties of NYT10 are used as unseen (for testing).
    
    \item \textbf{ACE05} is a corpus for event extraction, annotated by $33$ fine-grained types which are sub-types of $8$ coarse-grained main types such as  
    Life and Justice from the ACE (Automatic Content Extraction) ontology. Huang et al. \cite{huang2018zero} \rvc{make} two zero-shot event extraction settings: \textit{(i)} predicting $23$ unseen fine-grained sub-types by training on $1$, $3$, $5$, or $10$ seen sub-types; \textit{(ii)} predicting unseen sub-types that belong to other main types by training on seen sub-types of Justice.
\end{itemize}

\subsubsection{Text Classification}
Few-shot text classification is similar to zero-shot text classification: \rvc{the majority of the solutions mainly utilize different kinds of word embeddings, and the research on KG-aware method is rare. 
Rios et al. \cite{rios2018few} propose an ontology augmented CNN classifier for both few-shot and zero-shot text classification, while Sui et al. \cite{sui2021knowledge} utilize knowledge retrieved from the NELL KG for augmenting a network which calculates the matching of the input and the class.
Here are the benchmarks used in these two works:}
\begin{itemize}
\item \rvc{\textbf{MIMIC II} \cite{perotte2014diagnosis} and \textbf{MIMIC III} \cite{johnson2016mimic} are multi-label text classification benchmarks used in \cite{rios2018few}. Their labels are concepts in the ICD-9 \footnote{\url{https://www.cdc.gov/nchs/icd/icd9.htm}} ontology which is an international standard diagnostic classification for all general epidemiological, many health management purposes and clinical use.
MIMIC II has $18,822$ labels for training and $1,711$ labels for testing; while MIMIC III has $37,016$ labels for training and $1,356$ labels for testing.
}
\item \textbf{ARSC} is a \rvc{popular} benchmark for binary text classification of sentiment \cite{Blitzer2007BiographiesBB}, generated from Amazon reviews for $23$ products (classes). 
\rvc{In \cite{sui2021knowledge}, $12$ products including books, DVDs, electronics and kitchen appliances are selected as the unseen classes, for each of which $5$ labeled reviews are given, and NELL is used as the auxiliary KG.}
\end{itemize}

\subsubsection{Question Answering}

\rvc{Zero-shot and few-shot} question answering (QA)\footnote{The scope of QA is actually quite wide. It often includes or has a high overlap with quite a few problems such as VQA, Knowledge Base QA, Table QA, Machine Reading Comprehension (MRC). In this part, we just refer to the problem of giving an answer or answers to a natural language question w.r.t. a context described by text. 
}
started to attract wide attention in recent years, mainly due to the fast development of pre-trained language models such as BERT and GPT-3 which are inherently capable of addressing ZSL and FSL problems in NLP since a large quantity of knowledge are learned from large-scale corpora as parameters \cite{yang2021empirical,ma2021knowledge}. 
Similar to text classification, the output answer (class) is often regarded as an additional input and fed to a prediction model together with the question input. 
\rvc{
It is worth mentioning that the definition of zero-shot QA varies from paper to paper.  
Some are consistent with our general ZSL definition which mainly requests that the classes (answer labels) for prediction have no associated training data, while the others are not.
For example, Ma et al. \cite{ma2021knowledge} simply regard testing the model on a QA dataset that is different from the training QA datasets as zero-shot QA;
Wei et al. \cite{wei2021finetuned} fine-tune language models on a collection of datasets of some specific tasks (e.g., sentiment classification and summarization), test the models on datasets of different tasks (e.g., commonsen QA), and regard this as its zero-shot QA setting . 
}

Although pre-trained language models have contained much knowledge via large scale parameters, symbolic knowledge (including commonsense and domain knowledge with logics) from KGs are often complementary and beneficial for addressing zero-shot QA. Therefore, there have been some KG-aware zero-shot QA studies \cite{bosselut2021dynamic,banerjee2020self}.
For example, 
Banerjee et al. \cite{banerjee2020self} \rvc{model} the QA problem via knowledge triple learning where the context, question and answer are modeled as a triple, and the answer is predicted given the context and question.
Their knowledge triple learning model is learned from KG triples.
\cz{Similar to \cite{banerjee2020self}, Zhou et al. \cite{zhou2021encoding} also \rvc{frame} the multiple-choice QA task as a knowledge completion (triple prediction) problem, where the model is trained by alternatively masking the subjects and the objects in triples}.
Bosselut et al. \cite{bosselut2021dynamic} \rvc{use} COMET --- a Transformer-based model trained on commonsense KGs such as ConceptNet to generate a context-relevant commonsense triples for each QA sample, \rvc{and then infer the answer from these triples.

Similarly, KGs can also benefit few-shot QA.}
For example, Banerjee et al. \cite{banerjee2020self} directly \rvc{extend} their knowledge triple learning model from zero-shot QA to few-shot QA where $8\%$ of the training data are given as the few-shot samples.
Bosselut et al. \cite{bosselut2021dynamic} also \rvc{extend} their zero-shot QA method, which infers the answer of a question according to their context-relevant commonsense triples, to few-shot QA by using $4$, $10$ or $20$ validation samples in evaluation.
%
However, due to challenges such as retrieving exactly relevant knowledge from a large KG and injecting KG knowledge into pre-trained language models, the investigation of KG-aware zero-shot and few-shot QA is still quite preliminary.

\rvc{There have been a few widely used QA benchmarks such as PhysicalIQA for commonsense physical reasoning \cite{bisk2020piqa}}.
They can be used for benchmarking zero-shot and few-shot QA after \rvc{suitable dataset partitioning.}
We suggest benchmarks that are constructed with KGs or have been partially aligned with KG entities.
The tasks of these benchmarks often rely on external knowledge, and their corresponding external KGs can be directly used for evaluating KG-aware methods. 
Here are some such benchmarks: 
\begin{itemize}
\item \textbf{SocialIQa} \cite{sap2019social} is a large-scale QA resource to evaluates a model’s capability to understand the social dynamics underlying situations described in short text snippets. 
It has 38K \rvc{question-answer} pairs. Each sample consists of a context, a question about that context, and three multiple choice answers from crowdsourcing.
Commonsense knowledge (i.e., seeds for creating the contexts and answers) are extracted from an event KG \rvc{named} ATOMIC \cite{sap2019atomic}.
This dataset \rvc{is} used by Bosselut et al. \cite{bosselut2021dynamic} and Banerjee et al. \cite{banerjee2020self} for evaluating their KG-aware zero-shot and few-shot \rvc{QA} methods.

\item \textbf{CommonsenseQA} \cite{talmor2019commonsenseqa} is a challenging dataset for evaluating commonsense QA methods.
\rvc{In total, it has $12,247$ questions, each of which has} 
5 answer candidates. The ground truth answers are annotated by crowdsourcing
based on question relevant subgraphs of ConceptNet \cite{speer2017conceptnet}.
CommonsenseQA \rvc{is also} adopted by Banerjee et al. \cite{banerjee2020self} for evaluation.

\item \textbf{STORYCS} \cite{lucy2017distributional} consists of 5-sentence stories with annotated motivations and emotional responses. It is originally for emotion classification, where the labels are drawn from classical theories of psychology. Bosselut et al. \cite{bosselut2021dynamic} \rvc{transform} the classification task into a QA task by posing an individual question for each emotion label, and \rvc{use} it for evaluating their KG-aware method for both zero-shot and few-shot settings.


\item \textbf{aNLI} \cite{bhagavatula2020abductive}, \textbf{QASC} \cite{khot2020qasc}, \textbf{OpenBookQA} \cite{mihaylov2018can} and \textbf{ARC} \cite{bhakthavatsalam2021think} \rvc{are} adopted by Banerjee et al. \cite{banerjee2020self} for evaluating their KG-augmented triple learning model for zero-shot and few-shot QA, besides SocialIQa and CommonsenseQA.
Specifically, aNLI which has \rvc{171K question-answer pairs} is a dataset with commonsese knowledge, while QASC, OpenBookQA and ARC, whose sample sizes range from 6K to 10K, are three QA datasets with scientific knowledge.
\cz{\textbf{OpenBookQA} and \textbf{ARC} \rvc{are} also adopted by Zhou et al. \cite{zhou2021encoding} for zero-shot QA.}

\end{itemize}

\subsection{Knowledge Graph Completion}

KG completion is to infer \rvc{missing knowledge in a KG. Most existing studies aim at predicting} relational facts (triples), which is sometimes called link prediction.
In this \rvc{paper we mainly refer to these link prediction studies.
In a zero-shot or few-shot setting, we are required to handle  entities and/or relations that emerge} after the KG embeddings have been learned.
\rvc{Since 
the solutions to addressing unseen entities and unseen relations are quite different,
we introduce the studies for unseen entities and for} unseen relations separately.

\subsubsection{KG Completion with Unseen Entities}
\rvc{There have been quite a few methods on KG completion with unseen entities. They often utilize entities' auxiliary information such as name information, textual descriptions and attributes, following the} mapping-based paradigm \cite{shah2019open,hao2020inductive} and the class feature paradigm \cite{zhao2017zero,niu2021open,wang2021kepler,amador2021ontology,shi2018open,wang2021inductive,zha2021inductive,wang2021structure,yao2019kg}.
%
Various benchmarks have been proposed.
They are usually constructed based on \rvc{some existing} normal KG completion \rvc{benchmarks such as} FB15k \cite{bordes2013translating}, FB15k-237 \cite{toutanova2015representing}, WordNet11 \cite{socher2013reasoning}, WN18RR \cite{dettmers2018convolutional} \rvc{and NELL-995 \cite{xiong2017deeppath}, and some sub-KGs extracted from original KGs} such as DBpedia \cite{auer2007dbpedia} and Wikidata \cite{vrandevcic2014wikidata}.
Their entity auxiliary information is often collected from the benchmarks' original KGs or some associated public resources.
For example, the textual descriptions of entities in DBpedia50k, FB20k and Wikidata5M can be collected from DBpedia, Freebase and Wikipedia, respectively; \rvc{while the textual descriptions of entities of FB15k-237 in \cite{daza2021inductive}} are extracted from the introduction section of their corresponding Wikipedia pages.

\rvc{These benchmarks are often constructed following a common way.} 
Given an original KG completion benchmark, a set of entities are \rvc{first} selected as unseen entities. \rvc{Then} their associated triples in the training set are removed. \rvc{Next,} the relations that appear in both the training set and the testing set are \rvc{adopted, and the triples of the not adopted relations are removed in both the training set and the testing set}. 
For a testing triple to predict, there could be two cases: \textit{(i)} \rvc{its head (or tail)} is an unseen entity while the other is a seen entity, and \textit{(ii)} both its head and tail are unseen entities.
Accordingly, we regard the benchmark \rvc{with the first case testing triples \textit{semi-ZS}, the benchmark with the second case testing triples \textit{fully-ZS}, and the benchmark with both testing triples as \textit{mixture-ZS}.
Here are some typical benchmarks:}
\begin{itemize}
    \item \textbf{FB15k-237-OWE} \cite{shah2019open} is a typical \textit{semi-ZS} benchmark built on FB15k-237.
    First, testing triples whose tail entities are to be predicted are collected.
    Specifically, a set of tail entities are selected, and some associated head entities are randomly picked from the FB15k-237 triples (by uniform sampling over all the associated head entities).
    \rvc{Each picked head entity is removed from the training graph by moving all the triples whose heads are this entity to the testing set, and removing all the training triples whose tails are this entity.
   Testing triples whose head entities are to be predicted are processed in the same way.}
    Then a testing set is generated by merging the above two kinds of testing triples and removing the testing triples whose relations are not in the training set.
    This testing set is further splitted into a validation set and \rvc{a} final testing set. 
    The dataset contains $2,081$ unseen entities, $12,324$ seen entities and $235$ relations. The numbers of triples for training, validation and testing are $242,489$, $10,963$ and $36,250$, respectively. 
    
    \item \textbf{DBpedia50k} and \textbf{DBpedia500k} \cite{shi2018open} are also typical 
    \textit{semi-ZS} benchmarks, constructed in a similar way as FB15k-237-OWE. DBpedia50k has $49,900$ entities and $654$ relations, with $32,388$, $399$ and $10,969$ training, validation and testing triples, respectively.
   DBpedia500k has $517,475$ entities and $654$ relations, with $3,102,677$, $10,000$ and $1,155,937$ training, validation and testing triples.
    
    \item \textbf{Wikidata5M} \cite{wang2021kepler}, originally developed for evaluating text-aware KG embedding methods, is an important \textit{fully-ZS} benchmark.
    It is constructed based on the Wikidata dump and the English Wikipedia dump.
    Each entity in Wikidata is aligned to a Wikipedia page and this page's first section is extracted as the entity's textual description. 
    Entities with no Wikipedia pages or with descriptions being shorter than 5 words are discarded. 
    Next, all the relational facts (triples) are extracted from the Wikidata dump. One triple is kept if both of its entities are not discarded, and its relation has a corresponding nonempty page in Wikipedia. Otherwise, this triple is discarded.
    The benchmark contains $4,594,485$ entities, $822$ relations and $20,624,575$ triplets.
    To support the zero-shot setting, Wang et al. \cite{wang2021kepler} randomly \rvc{extract} two sub-KGs \rvc{as the validation set and the testing set,
    and use the remaining as the training set. 
    The three sets, respectively,} have $4,579,609$, $7,374$ and $7,475$ entities, $822$, $199$ and $201$ relations, and $20,496,514$, $6,699$ and $6,894$ triples.
    
    \item \textbf{FB20k} \cite{xie2016representation} is a  \rvc{benchmark whose testing triples may involve unseen entities.} It has the same training set and validation set as the normal KG completion \rvc{benchmark} FB15k, but extends FB15k's testing set by adding triples involving unseen entities.
    Specifically, a candidate set of unseen entities are first selected from Freebase. They should be associated with some entities in FB15k entities within one hop.
    \rvc{Then some new triples whose relations are ensured to be already in FB15k are extracted from Freebase and added to the testing set.
    These new testing triples have four kinds:} those whose head and tail are both seen entities, those whose heads are unseen and whose tails are seen, those whose tails are unseen and whose heads are seen, and those whose heads and tails are both unseen.
    The first kind of testing triples are for normal KG completion, while the other three kinds are for zero-shot KG completion.
    So the task of FB20k can be understood as generalized zero-shot KG completion.
    The numbers of the test triples of the above four types are $57,803$, $18,753$, $11,586$, and $151$, respectively, and all these triples involve $19,923$ entities.
    The subsets of FB15k-237 and WN18RR proposed in \cite{daza2021inductive} are \rvc{similar.}
\end{itemize}

In few-shot KG completion, unseen entities usually have a small number of associated triples \rvc{given}. The current methods often aim to fully utilize these triples, mainly \rvc{following} the propagation-based paradigm \cite{zhao2020attention,albooyeh2020out,hamaguchi2017knowledge,bhowmik2020explainable,wang2019logic,dai2020inductively,ali2021improving}, the transfer-based paradigm \cite{teru2020inductive,liu2021indigo,chen2021topology,sadeghian2019drum} and the optimization-based paradigm \cite{baek2020learning}.
%
\rvc{Several few-shot KG completion benchmarks with unseen entities have also been constructed based on normal} KG completion benchmarks.

According to the type of the entity that an unseen entity is linked to, we categorize these benchmarks into three categories.
\rvc{For the first category, the entity linked to is seen in training. Thus the}
few-shot triples can be utilized to propagate embeddings from seen entities to unseen entities by e.g., GNNs \cite{hamaguchi2017knowledge,wang2019logic,bhowmik2020explainable}.
Typical benchmarks of this category include subsets extracted from WordNet11 by \cite{hamaguchi2017knowledge}, subsets extracted from FB15k by \cite{wang2019logic}, and subsets extracted from WN18RR, FB15k-237 and NELL-995 by \cite{bhowmik2020explainable}.
For the second category, the entity linked to is also an unseen entity.
These benchmarks are to evaluate the generalization ability of a model trained on one KG to \rvc{another KG or an emerging sub-KG with different entities.
Methods of the transfer-based paradigm are often be adopted.} 
Typical benchmarks of this category include subsets of WN18RR, FB15k-237 and NELL-995 extracted by \cite{teru2020inductive}.
In the third category, the entity linked to can be either unseen or seen.
Typical benchmarks include subsets of WN18RR, FB15k-237 and NELL-995 contributed in \cite{baek2020learning} where a meta learning method is \rvc{often} applied to learn embeddings of the unseen entities from their few-shot triples.
Next, we will introduce more details of some representative benchmarks of each category:

\begin{itemize}
    \item \textbf{Subsets of WordNet11 by Hamaguchi et al.} \cite{hamaguchi2017knowledge} are of the first category. \rvc{They are constructed from the normal KG completion benchmark WordNet11} in the following way.
    First, entities in the original testing set are extracted as unseen entities, while all the other entities are regarded as seen. Among \rvc{the} unseen entities, those that are associated to only seen entities in the original training triples are kept and the \rvc{others} are discarded. 
    Second, the original training triples that do not contain any unseen entities are selected for a new training set, those that contain exactly one unseen entity are selected as few-shot samples, and those that contain two unseen entities are discarded.
    Next, a new testing set is constructed \rvc{from} the original testing triples \rvc{by} removing those containing no unseen entities.
    Nine subsets of different scales are extracted for the few-shot setting, by setting the size of testing triples for extracting unseen entities to $1,000$, $3,000$ and $5,000$, and by setting the position for extracting unseen entities to head, tail and both.

    \item \textbf{Subsets of WN18RR, FB15k-237 and NELL-995 by Teru et al.} \cite{teru2020inductive} are of the second category.
    They are constructed in the following way.
    Given one original benchmark, two disjoint graphs are sampled:
    \rvc{\textit{train-graph} for training and \textit{ind-test-graph} for testing.
    It is ensured that their entity sets are disjoint, while the relations of \textit{ind-test-graph} are all involved in \textit{train-graph}}.
    In particular, $10\%$ of the triples of the \textit{ind-test-graph} are randomly selected for testing.
    These benchmarks are also adopted for evaluation in \cite{chen2021topology}.
    
    \item \textbf{Subsets of WN18RR, FB15k-237 and NELL-995 by Baek et al.} \cite{baek2020learning} \rvc{are of} the third category.
    These subsets are extracted from each original \rvc{benchmark as follows.} 
    First, a set of entities, which have a relatively small amount of associated triples, are randomly sampled as the unseen entities,
    and they are further partitioned and used for constructing three meta sets of triples: a meta-training set, a meta-validation set and a meta-testing sets.
    The other entities in the original benchmark are regarded as seen entities.
    Second, triples composed of seen entities alone are extracted to construct a graph named \textit{In-Graph}.
    Finally, the meta sets are cleaned, such that each of their triples has at least one unseen entity and all the triples are out of \textit{In-Graph}.
\end{itemize}

\subsubsection{KG Completion with Unseen Relations}

Zero-shot KG completion with unseen relations usually utilize the \rvc{relations'} auxiliary information such as their names and descriptions, mainly \rvc{following} the data augmentation paradigm \cite{geng2021ontozsl,qin2020generative,geng2023benchmarking} and the class feature paradigm \cite{zha2021inductive,wang2021structure,yao2019kg};
while few-shot KG completion with unseen relations usually relies on the few-shot triples using methods of the optimization-based paradigm \cite{wang2019meta,chen2019meta,zhang2020fewb,lv2019adapting}, the mapping-based paradigm \cite{zhang2020fewa,xiong2018one} and the propagation-based paradigm \cite{zhao2020attention}.
\rvc{In comparison with} KG completion with unseen entities, there are fewer benchmarks for KG completion with unseen relations.
We find NELL-ZS and Wiki-ZS for the zero-shot setting, and NELL-One and Wiki-One for the few-shot setting.
NELL-ZS and NELL-One are sub-KGs extracted from NELL, while Wiki-ZS and Wiki-One are sub-KGs extracted from Wikidata.
Their details are introduced as follows:

\begin{itemize}
    \item \textbf{NELL-ZS} and  \textbf{Wiki-ZS} \cite{qin2020generative} 
    \rvc{both have a training set with triples of seen relations, a validation set and a testing set with triples of unseen relations.}
    The entities in the testing triples and the validation triples have all been involved in some training triples.
    NELL-ZS has $139$, $10$ and $32$ training, validation and testing relations, \rvc{respectively,} and $65,567$ entities; while Wiki-ZS has $469$, $20$, $48$ training, validation and testing relations, respectively, and $605,812$ entities.
    \rvc{For both NELL-ZS and Wiki-ZS, Qin et al. \cite{qin2020generative} use relation textual descriptions as the auxiliary information, while}
    Geng et al. \cite{geng2021ontozsl,geng2023benchmarking} \rvc{construct} ontological schemas, which contain not only textual information but also relation hierarchies, relation domains and ranges, relation characteristics and so on.
    
    \item \textbf{NELL-One} and \textbf{Wiki-One} \rvc{are} originally developed by Xiong et al. \cite{xiong2018one} for evaluating \rvc{one-shot} KG completion with unseen relations.
    In construction, relations that are associated with less than $500$ triples but more than $50$ are extracted from the original KGs as task relations (i.e., one relation corresponds to one task). 
    In NELL-One, $67$ such relations are extracted and they are partitioned into $51$, $5$ and $11$ for constructing triples of \rvc{the training, validation and testing set}, respectively; while in Wiki-One, $183$ such relations are extracted and partitioned into $133$, $16$ and $34$ for constructing triples of \rvc{the training, validation and testing set}, respectively.
    $68,545$ entities are extracted for NELL-One, and $4,838,244$ entities are extracted for Wiki-One.
    In addition, another $291$ and $639$ relations are extracted, respectively, as background relations constructing more triples for the entities.
    Note that these two benchmarks can also be simply revised \rvc{for more general few-shot KG completion by adding more than one triples.}
\end{itemize}

\section{
\rvc{Open Problems}
}\label{sec:challenge}


\subsection{\rvc{Knowledge Graph Quality}
}
\rvc{For a ZSL or FSL task, one critical challenge is constructing a customized KG with exactly necessary and high-quality knowledge.
Although we now can re-use existing KGs, extract knowledge from some task data, and curate knowledge with domain experts, some open problems still remain.}

First, \rvc{knowledge and data integration is rarely investigated, and the impact of low-quality knowledge, which could be biased or even erroneous, has not been studied in the current KG-aware ZSL and FSL works}.
%
Second, the coverage of necessary knowledge and the ratio of irrelevant knowledge are often ignored in investigating a KG's usefulness towards a ZSL or FSL task, and there is a shortage of methods that are able to \rvc{(semi-)automatically} retrieve relevant knowledge from a large scale KG for a given task.
%
\rvc{Third, several popular knowledge sources such as natural language text, Web tables and databases, and their corresponding knowledge extraction methods like open information extraction, have been rarely explored for constructing the auxiliary KG for ZSL or FSL. This could be a promising direction to improve the coverage and quality of the auxiliary KG.
Fourth crowdsourcing and human-on-the-loop techniques for data curation and KG construction (e.g., \cite{jiang2018knowledge}) are also worth investigating as another potential way for further augmenting ZSL and FSL.}

\subsection{\rvc{Learning Paradigms}}

\subsubsection{\rvc{KG-aware ZSL}}
The mapping-based paradigm \rvc{has been} widely investigated, while the data augmentation paradigm has only four methods, \rvc{one of which uses rules while the remaining three of which use generation models.
Mapping-based methods are often biased to unseen classes in prediction, while the generation-based paradigm can flexible choose the model and avoid the bias after data are generated.
Thus we think generation-based methods conditioned on KG embeddings is worth of more investigation in the future.
Meanwhile, it is hard to use rules to generate numeric samples or features, but in KG completion, it is feasible to use ontological schemas and logical rules to infer triples for the unseen entities and/or relations. It would be a promising direction to combine symbolic reasoning with data augmentation.}

%
\rvc{Regarding the propagation-based paradigm, belief propagation has not been widely investigated, but would be a good solution for some CV tasks such as scene graph extraction and VQA, where multiple objects in an image or video, as well as their semantic relationships, need to be recognized.}
%
\rvc{With the development of pre-trained language models, the class feature paradigm is becoming more and more popular, especially for tasks whose inputs are text, such as text classification, question answering and knowledge extraction.
We think this trend will continue in the future, while KGs will still play an important role by providing symbolic knowledge that these parameter-based language models cannot represent.}

\rvc{
\subsubsection{KG-aware FSL}
Many FSL methods focus on utilizing the few-shot samples via applying meta learning algorithms or extending the ZSL methods.
It is still challenging to combine the KG auxiliary information and the few-shot samples. 
For the data augmentation paradigm, how to merge the generated samples and the few-shot samples? 
For the optimization-based paradigm, how to use KG to guide the meta learning algorithm? 
For the transfer-based paradigm, how to guide the model transfer with KG?
For embedding propagation for KG completion, how to augment the propagation models such as GNN with auxiliary information especially the ontological schema?
We think all these are still open problems and worth further investigation in the future.
}

\subsection{\rvc{Zero-shot and Few-shot Learning in} KG Construction}
Nowadays KG construction uses not only heuristics (e.g., hand-craft rules and templates), symbolic knowledge engineering and manual curation, but also machine learning prediction for (semi-)automation \cite{weikum2020machine}.
\rvc{Prediction tasks range from knowledge extraction from different data sources such as text, tables and Web pages, to knowledge curation such as KG completion, entity alignment, entity resolution, entity typing and schema inference. Many such tasks rely on supervised learning, but often suffer from the shortage of labeled samples.
Although some tasks such as entity linking and KG completion have been widely investigated in FSL and ZSL, developing robust ZSL and FSL methods for these prediction tasks under a KG context is still an open problem and should attract wider attention. 
Meanwhile, some KG construction and curation tasks, such as entity typing and table to KG matching, some dynamic KG contexts with e.g., involving schema and/or data, and some complex knowledge representations such as OWL ontology and Datalog rule, can be considered for new benchmarks for KG-aware ZSL and FSL.}

\rvc{
\subsection{Benchmarking}
Although there have been some benchmarking studies for ZSL and FSL \cite{xian2018zero,yin2019benchmarking}, systematic evaluation and comparison of KG-aware methods is still not enough.
The existing KG-aware ZSL and FSL benchmarks are usually associated with fixed KGs. The current methodology studies do not consider the impact of different settings on knowledge coverage, representation and quality settings, and rarely apply one method to different tasks, which cannot show the generalization capability.
Our recent benchmarking work \cite{geng2023benchmarking} has analyzed the impact of different KG semantics such as textual information, attributes and RDFS schemas on two typical and representative ZSL methods OntoZSL \cite{geng2021ontozsl} and DeViSE \cite{frome2013devise} for three tasks (see results in the original paper), but fare and comprehensive comparison of more methods across different datasets, different knowledge settings and different tasks are urgently needed in the future.
Meanwhile, more benchmarks should be developed to cover more domains where KGs are widely used such as health science domains.
}

\section{Conclusion}\label{sec:conclusion}

\rvc{KGs have become popular auxiliary information for augmenting ZSL and FSL, and at the same time KG construction also involve many prediction tasks with zero-shot and few-shot settings.}
Thus \hy{KG-aware ZSL and FSL have gained widespread attention and popularity in many domains such as CV, NLP, ML and the semantic Web}.
In this survey, we systematically \rvc{review} over $90$ KG-aware studies for addressing \rvc{sample shortage in ML} from perspectives of the KG, the methodology and the application.
\rvc{
The content covers \textit{(i)} the introduction of KGs that have been applied  and the methods for constructing such task-specific KGs, \textit{(ii)} the review of the KG-aware ZSL and FSL methods of each paradigm, and \textit{(iii)} the presentation of the development of ZSL and FSL research for different tasks in CV, NLP and KG completion, as well as the resources that can be used for evaluating KG-aware ZSL and FSL methods.}
Besides, we \rvc{have} also analyzed and discussed  the challenges of KG-aware \rvc{zero-shot anf few-shot learning, and some potential future directions.}

\section*{Acknowledgments}
This work \rvc{is} supported by 
eBay, Samsung Research UK 
and the EPSRC projects ConCur (EP/V050869/1) and UK FIRES (EP/S019111/1).

\balance
\bibliographystyle{IEEEtranN}
\bibliography{sample-base}

\end{document}